%% file: main.tex
\documentclass[conference]{IEEEtran}
\IEEEoverridecommandlockouts
\usepackage[a-1b]{pdfx}
\usepackage{balance}
\usepackage{hyperref} 
\usepackage{booktabs} 
\usepackage{xcolor}    
\usepackage{colortbl} 
\usepackage{subcaption}
\usepackage{multirow}

\usepackage{tikz}
\usetikzlibrary{arrows.meta,positioning,fit,shapes.geometric}

\usepackage{cite}
\usepackage{amsmath,amssymb,amsfonts}
\usepackage{algorithmic}
\usepackage{graphicx}
\usepackage{textcomp}
\usepackage{tabularx}  
\usepackage[numbers]{natbib}  

\usepackage[capitalise]{cleveref} % for \Cref/\cref

\def\BibTeX{{\rm B\kern-.05em{\sc i\kern-.025em b}\kern-.08em
    T\kern-.1667em\lower.7ex\hbox{E}\kern-.125emX}}

\makeatletter
\def\url@leostyle{%
  \renewcommand{\UrlFont}{\small\normalfont} % small, normal font
  \renewcommand\url[1]{}% suppress printing of URLs completely
}
\makeatother
\makeatletter
\renewcommand\thebibliography[1]{%
  \section*{\refname}%
  \small % reduce font size
  \list{\@biblabel{\@arabic\c@enumiv}}%
       {\settowidth\labelwidth{\@biblabel{#1}}%
        \leftmargin\labelwidth
        \advance\leftmargin\labelsep
        \itemsep=0pt
        \parsep=0pt
        \rightmargin=0pt % makes sure it's justified to the right
        \listparindent=0pt
        \itemindent=0pt
        \topsep=0pt
        \partopsep=0pt
       }%
  \def\newblock{\hskip .11em plus .33em minus .07em}%
  \sloppy\clubpenalty4000\widowpenalty4000%
  \sfcode`\.=\@m}
\makeatother
\begin{document}

% \makeatletter
% \def\@IEEEtitlefont{\normalfont\bfseries\Large}
% \makeatother

\title{Learning to Land Anywhere: Transferable Generative Models for Aircraft Trajectories}

% \title{{\fontsize{22pt}{18pt}\selectfont
% Leveraging Learned Representations from Generative Models for Analyzing ATM Operational Patterns}}

\author{
    \IEEEauthorblockN{Olav Finne Præsteng Larsen$^{2}$, Massimiliano Ruocco$^{1,2}$, Michail Spitieris$^{1}$, Abdulmajid Murad$^{1}$, Martina Ragosta$^{1}$}
    \IEEEauthorblockA{
        \parbox{0.5\textwidth}{
            \centering
            $^1$Department of Software Engineering, Safety and Security \\ 
            SINTEF Digital \\
            Trondheim, Norway \\
            \{abdulmajid.murad, massimiliano.ruocco, michail.spitieris, martina.ragosta\}@sintef.no
        }
        \hspace{0.05\textwidth}
        \parbox{0.4\textwidth}{
            \centering
            $^2$Department of Computer Science \\
            Norwegian University of Science and Technology \\
            Trondheim, Norway \\
            massimiliano.ruocco@ntnu.no
        }
    }
}

\maketitle

\begin{abstract}
\input{sections/abstract}
\end{abstract}


\input{sections/introduction}
\input{sections/sota}

\input{sections/method}
\input{sections/experimentalsetting}

\input{sections/results}
\input{sections/conclusion}

\section*{Acknowledgment}
This paper is based on research conducted within the SynthAIr project, which has received funding from the SESAR Joint Undertaking under the European Union’s Horizon Europe research and innovation program (grant agreement No. 101114847). %The views and opinions expressed in this paper are solely those of the authors and do not necessarily reflect those of the European Union or the SESAR 3 Joint Undertaking. Neither the European Union nor the SESAR 3 Joint Undertaking can be held responsible for any use of the information contained herein.

\balance

\bibliographystyle{IEEEtran_noonline}

\bibliography{references_ok}

\end{document}

%% file: sections/abstract.tex
Access to trajectory data is a key requirement for developing and validating Air Traffic Management (ATM) solutions, yet many secondary and regional airports face severe data scarcity. This limits the applicability of machine learning methods and the ability to perform large-scale simulations or ``what-if'' analyses. In this paper, we investigate whether generative models trained on data-rich airports can be efficiently adapted to data-scarce airports using transfer learning. We adapt state-of-the-art diffusion- and flow-matching--based architectures to the aviation domain and evaluate their transferability between Zürich (source) and Dublin (target) landing trajectory datasets. Models are pretrained on Zürich and fine-tuned on Dublin with varying amounts of local data, ranging from 0\% to 100\%. Results show that diffusion-based models achieve competitive performance with as little as 5\% of the Dublin data and reach baseline-level performance around 20\%, consistently outperforming models trained from scratch across metrics and visual inspections. Latent Flow Matching and Latent Diffusion models also benefit from pretraining, though with more variable gains, while Flow Matching models show weaker generalization. Despite challenges in capturing rare trajectory patterns, these findings demonstrate the potential of transfer learning to substantially reduce data requirements for trajectory generation in ATM, enabling realistic synthetic data generation even in environments with limited historical records.

%Trajectory data is essential for developing Air Traffic Management (ATM) solutions, yet secondary airports often suffer from severe data scarcity. We address this challenge by investigating whether generative models trained on data-rich airports can be transferred to data-scarce ones through transfer learning. Using landing trajectory datasets from Zürich (source) and Dublin (target), we evaluate diffusion, flow-matching, and latent generative models across data splits from 0\% to 100\% of the target dataset. Results show that diffusion-based models transfer particularly well: with only 5\% of Dublin data, they already achieve competitive performance, and at 20\% they match or surpass models trained from scratch on the full dataset. Latent models also benefit from pretraining, while flow-matching shows weaker gains. These findings highlight transfer learning as a practical strategy to reduce data requirements for trajectory generation, enabling realistic synthetic data in environments where historical records are limited.

%% file: sections/introduction.tex
\section{Introduction}
The development of robust Air Traffic Management (ATM) systems increasingly relies on the availability of high-quality trajectory data. Such data underpins the design of machine learning algorithms, the validation of new operational procedures, and large-scale simulation studies. However, while major hub airports typically record large volumes of flight data, many secondary and regional airports suffer from severe data scarcity. This imbalance limits the ability to test and validate new ATM solutions across diverse operational contexts. Moreover, the exploration of rare events, edge cases, and ``what-if'' scenarios requires datasets far larger and more varied than what is available from real-world operations.  

Recent advances in generative modeling have shown promise in addressing this challenge by synthesizing realistic flight trajectories. Techniques such as Variational Autoencoders (VAEs), Generative Adversarial Networks (GANs), and Diffusion Models have demonstrated their ability to capture the distributional characteristics of real flight data and generate plausible synthetic samples. Nevertheless, these approaches typically assume access to a sufficient amount of local training data. In practice, this assumption often fails for airports with limited operations or incomplete historical records, restricting the applicability of generative methods where they are needed most.  

In this paper, we investigate whether transfer learning can mitigate these challenges. Specifically, we explore whether generative models trained on a data-rich airport can be effectively adapted to airports with scarce data. By pretraining on a large landing trajectory dataset from Zürich Airport and fine-tuning on limited data from Dublin Airport, we evaluate the transferability of state-of-the-art generative architectures, including Diffusion Models, Flow Matching, and Latent Flow Matching.  

The main contributions of this work are as follows:
\begin{itemize}
    %\item We introduce introduce of novel models LDM and LFM
    \item We present, to the best of our knowledge, the first systematic study of transfer learning for generative trajectory models in the ATM domain.  
    \item We demonstrate that diffusion-based models can achieve competitive performance with as little as 5\% of the target airport’s data and reach baseline-level performance around 20\%, significantly reducing local data requirements.  
    \item We provide a comparative analysis of different generative approaches, highlighting the strengths of diffusion-based architectures and the limitations of flow-matching methods for transfer learning.  
    \item We discuss the implications of transfer learning for enabling realistic trajectory generation in data-scarce environments, thereby broadening the applicability of synthetic data in ATM research and development.  
\end{itemize}

By showing that generative models can transfer knowledge across airports, this work paves the way toward scalable synthetic trajectory generation, supporting the development of data-driven ATM systems even in operational contexts where historical records are limited.  

%% file: sections/sota.tex
\section{State of the Art}
\label{sota}

Research on synthetic flight landing trajectories lies at the intersection of deep generative modeling and aviation operations. Progress in deep generative models has been rapid, spanning Variational Autoencoders (VAE) \cite{kingma_auto-encoding_2022}, Generative Adversarial Network (GAN) \cite{goodfellow_generative_2014}, Diffusion Models (DM) \cite{sohl-dickstein_deep_2015,ho_denoising_2020,song_denoising_2022}, and, most recently, Flow Matching (FM) \cite{lipman_flow_2023}. Latent variants such as Latent Diffusion Models (LDM) \cite{rombach_high-resolution_2021} and latent flow matching (LFM) \cite{dao_flow_2023} combine strong sample quality with improved efficiency by shifting generation to a compressed space. While most architectural innovation has emerged in computer vision, a growing body of work adapts these ideas to spatio-temporal trajectories, where temporal dependencies, multi-modality, and physical constraints are central. In particular this work focus primarily on advances in trajectory generation in the ATM domain. This is currently a growing research area aiming to address challenges related to data accessibility, privacy, and the complexity of flight data \cite{eurocontrol_fly_2020}. 

Aircraft trajectory generation can be formulated as the challenge of generating a set of fully timestamped coordinates for an aircraft that reflect physical behavior and the statistical properties of the spatio-temporal bounds \cite{olive_framework_2021}. The existing literature on generative flight trajectory models divides them into two categories: model-driven and data-driven approaches \cite{olive_framework_2021, wijnands_generation_2024, murad_synthetic_2025}. Model-driven approaches rely on mathematical models of flight, ensuring that the trajectories follow the laws of aerodynamics and other operational constraints \cite{olive_framework_2021}. %Examples include the Base of Aircraft Data (BADA) model \cite{nuic_bada_2010} and OpenAP \cite{sun_openap_2020}, which simulate trajectories based on aircraft performance parameters and operational rules. These methods generate physically accurate and realistic flight paths but are limited in capturing the stochastic variations and diverse patterns present in real-world trajectories \cite{olive_framework_2021}.

Data-driven methods use machine learning techniques to model and generate flight trajectories directly from data \cite{olive_framework_2021}. These methods, such as TimeGAN \cite{wijnands_generation_2024}, TCVAE and VampPrior\cite{krauth_deep_2023}, have shown promise in capturing temporal dependencies and generating realistic landing trajectories and \cite{murad_synthetic_2025} that applied the proposed TimeVQVAE architecture \cite{lee_vector_2023} to generate complete flight trajectories and introduced new evaluation methods tailored to trajectory generation Interestingly, \cite{krauth_deep_2023} generates trajectories by groundspeed and track. In contrast, \cite{wijnands_generation_2024} uses longitude and latitude, which is the most common approach in other domains, demonstrating that there are different approaches to modeling the data to be learned.

While the application of data-driven generative models in aviation remains underdeveloped, significant advances have been made in other fields, particularly urban traffic, where state-of-the-art generative models such as VAEs \cite{chen_trajvae_2021}, GANs \cite{yoon_time-series_2019}, and DMs \cite{zhu_difftraj_2023, zhu_controltraj_2024} have successfully learned complex trajectory data distributions. Recent advancements in DMs \cite{zhu_difftraj_2023, kong_diffwave_2021, wei_diff-rntraj_2024} have demonstrated significant potential for spatio-temporal data generation. Although aviation and urban traffic differ due to the three-dimensional and sparse nature of airspaces compared to the strict topological constraints of urban traffic \cite{zhu_controltraj_2024}, the DMs’ ability to handle complex, high-dimensional distributions \cite{song_denoising_2022, ho_denoising_2020} suggests they can overcome the challenges posed by flight trajectories. These are challenges such as the need for physical realism, airspace restrictions, and various weather conditions \cite{olive_framework_2021, gultepe_review_2019}.

While newer DMs have been proposed in urban traffic \cite{zhu_controltraj_2024, wei_diff-rntraj_2024}, these are specifically tailored to the urban traffic domain by incorporating road constraint information into the models to improve performance. Therefore, DiffTraj \cite{zhu_difftraj_2023} is the chosen diffusion architecture to be adapted to the aviation domain, as it has good performance while not implementing strict domain-specific constraints, potentially making it well-suited for generating flight trajectories. Additionally, newer advancements have been made in general deep generative models with the introduction of FM models \cite{lipman_flow_2023}, an alternative to DMs. 

The use of DMs and FMs also opens up possibilities for other methodologies, such as latent diffusion models (LDM) \cite{rombach_high-resolution_2021} and latent flow matching (LFM) \cite{dao_flow_2023}, which are unexplored in the field of flight trajectory generation. The proven TCVAE from \cite{krauth_deep_2023} can be used to represent trajectories in a latent space, and the DiffTraj architecture can then be used to generate the latent representations to be decoded into trajectories. 

Finally, while diffusion has also been explored for \emph{prediction} in terminal airspace \cite{yin_aircraft_2025}, our emphasis remains on \emph{generation} rather than prediction, i.e., learning the underlying distribution to produce diverse, plausible samples \cite{olive_framework_2021,murphy_probabilistic_2023}.

Building on prior work in deep generative model, our work focus on the underexplored question of transfer learning: we systematically test whether state-of-the-art deep generative models pretrained on a data-rich source can be efficiently fine-tuned on a data-scarce target to match or surpass target-only baselines.

%% file: sections/method.tex
\section{Model Architectures}

We investigate four families of generative models for aircraft landing trajectories: 
\emph{Diffusion Models (DMs)}, \emph{Flow Matching (FM)}, \emph{Latent Diffusion Models (LDMs)}, 
and \emph{Latent Flow Matching Models (LFMs)}. All architectures are adapted from recent advances in spatio-temporal generative modeling and extended to the aviation domain.  

\subsection{Diffusion Models (DMs)}  
Diffusion probabilistic models~\cite{sohl-dickstein_deep_2015,ho_denoising_2020,song_denoising_2022} construct a forward process that gradually corrupts data with Gaussian noise, and a reverse process trained to denoise step by step. For trajectory synthesis, we adapt the \emph{DiffTraj} architecture~\cite{zhu_difftraj_2023}, which employs a UNet backbone with residual and attention blocks, together with wide-and-deep embeddings for conditional inputs (airport/runway). Unlike urban traffic where strict road-network constraints can be enforced, aviation requires flexibility in sparse three-dimensional airspaces. By removing road-specific constraints while retaining the robust denoising framework, this work represents one of the first systematic applications of diffusion models to aircraft landing trajectories.  

\subsection{Flow Matching (FM)}  
Flow Matching (FM)~\cite{lipman_flow_2023} is a recently introduced generative paradigm that treats synthesis as a continuous-time transport problem, learning velocity fields that map a simple prior distribution to the target data distribution. Compared to diffusion, FM eliminates the need for discrete noise schedules and sampling schedulers, potentially offering faster generation. To the best of our knowledge, this is the \emph{first study to apply FM to trajectory generation in aviation}, extending a cutting-edge generative approach from computer vision into ATM research.  

\subsection{Latent Variants (LDM and LFM)}  
Both diffusion and FM models are combined with a Temporal Convolutional Variational Autoencoder (TCVAE)~\cite{krauth_deep_2023}, which compresses trajectories into a structured latent representation. Operating in this latent space reduces dimensionality, accelerates training and sampling, and enables more compact generative modeling.  

In the \emph{Latent Diffusion Model (LDM)}~\cite{rombach_high-resolution_2021}, the denoising process occurs in latent space, and the decoded outputs preserve the temporal dependencies of flight dynamics.  
In the \emph{Latent Flow Matching (LFM)}~\cite{dao_flow_2023}, the continuous transport dynamics are applied directly to latent embeddings, allowing FM to exploit compact trajectory representations.  

To our knowledge, this work represents the \emph{first use of LDMs and LFMs for flight trajectory generation}, contributing novel architectures to the ATM domain.  

\subsection{Conditional Generation and Representation}  
All architectures are conditioned on airport identity. At Zürich (LSZH), runway identifiers are available, while Dublin (EIDW) is represented by a single airport token due to missing runway labels. To ensure airport-agnostic generalization and avoid geographic leakage, we train primarily in a kinematic representation—track, groundspeed, altitude, and elapsed time—rather than raw latitude–longitude. This design choice avoids distortions from map projections and improves transferability across airports.  

%\subsection{Relevance for Transfer Learning}  
%These four architectures form the basis for our transfer-learning study, in which models are pretrained on a data-rich source airport (Zürich, LSZH) and subsequently fine-tuned on a data-scarce target airport (Dublin, EIDW) under varying amounts of available local data. By comparing pretrained models against from-scratch baselines, we systematically evaluate how architectural choices impact cross-airport transferability.  

%% file: sections/experimentalsetting.tex
\section{Experimental Settings}
These four architectures form the basis for our transfer-learning study, in which models are pretrained on a data-rich source airport (Zürich, LSZH) and subsequently fine-tuned on a data-scarce target airport (Dublin, EIDW) under varying amounts of available local data. By comparing pretrained models against from-scratch baselines, we systematically evaluate how architectural choices impact cross-airport transferability. 

\subsection{Dataset}
\label{subsec:datasets}

%Our study focus on whether generative models trained on a data-rich airport can be adapted to a data-scarce airport with minimal local data. Concretely, 
We use publicly available \emph{landing-trajectory} datasets for Zürich (LSZH) and Dublin (EIDW) from the \texttt{traffic} library \citep{olive_traffic_2019}, originally derived from OpenSky \citep{schafer_bringing_2014}. LSZH is employed as the data-rich \emph{source} domain due to its widespread use in prior trajectory-generation studies \citep{wijnands_generation_2024,krauth_deep_2023}; EIDW serves as the \emph{target} domain with comparable scale. To ensure comparability and computational efficiency, all trajectories are resampled to a fixed length of $T{=}200$ time steps using \texttt{traffic}, following established practice for sequence models \citep{zhu_difftraj_2023}. Details about the data are summarized in Table \ref{tab:landing_datasets}.

\begin{table}[ht]
  \centering
  \caption{Landing datasets used in this study.}
  \label{tab:landing_datasets}
  \begin{tabular}{lcccc}
    \toprule
    \textbf{Airport} & \textbf{Period} & \textbf{Samples} & \textbf{Runways} & \textbf{ICAO} \\
    \midrule
    Z\"urich & 2019-10\,--\,2019-11 & $\sim$19{,}000 & 4 & LSZH \\
    Dublin  & 2019-10\,--\,2019-11 & $\sim$20{,}000 & 4 & EIDW \\
    \bottomrule
  \end{tabular}
\end{table}

All datasets are standardized, with \emph{separate} scalers for trajectory variables and conditional inputs. %scalers are fit on the training split and applied to validation/test to avoid leakage. 
For EIDW, runway labels are unavailable; during conditioning. We therefore use a single airport token (``EIDW'') instead of the runway identity, while LSZH includes runway-specific labels. %Train/validation/test splits follow the protocol in \autoref{cha:ResearchAndResults:Eval} and are kept identical across model families to isolate the effects of pretraining and data fraction.

%\subsection{Experimental Plan and Setup}

\subsection{Evaluation Metrics}
Evaluating generative models for flight trajectories is a nontrivial task: no single metric jointly captures statistical fidelity, geometric plausibility, and operational realism. Existing frameworks range from expert assessments and physics-aware simulation (e.g., BlueSky simulator \citep{hoekstra_bluesky_2016}) to compact statistical criteria \citep{olive_framework_2021}. The former offer operational insight but are resource intensive; the latter support reproducible model selection but do not certify operational validity.

Given our focus on transfer learning and comparative analysis across models and data budgets, we adopt a lightweight quantitative protocol. Following prior work \citep{wijnands_generation_2024,krauth_deep_2023}, metrics are computed in the lateral plane using longitude and latitude only—excluding altitude—to emphasize spatial alignment. When models are trained in a kinematic representation (track and groundspeed), geographic footprints are reconstructed under the same convention for evaluation.

We assess \emph{similarity} to quantify distributional alignment between real and synthetic trajectories using quantitative, reproducible metrics that capture both statistical and structural properties. As in \citep{wijnands_generation_2024,krauth_deep_2023}, we report energy distance (e-distance), a Euclidean-distance–based measure of distributional discrepancy \citep{szekely_testing_2004}, together with Maximum Mean Discrepancy (MMD) \citep{viehmann_partial_2021} and Dynamic Time Warping (DTW) \citep{olive_framework_2021} to capture, respectively, global distributional differences and local shape-level deviations. Where informative, we also compute Kullback–Leibler divergence (KL) and Jensen–Shannon divergence (JSD), inspired by urban-traffic evaluation \citep{zhu_difftraj_2023}, which are effective for comparing spatial density maps (trajectory heatmaps) and provide a complementary view of how well the synthetic distribution approximates the real one.

We evaluate \emph{diversity} via dimensionality-reduction diagnostics. Principal Component Analysis (PCA) and t-distributed stochastic neighbor embedding (t-SNE) visualize coverage, cluster structure, and outliers by projecting real and synthetic sets into a low-dimensional space. Following \citep{yoon_time-series_2019}, trajectories are flattened along the temporal dimension before projection to enable a like-for-like comparison of distributional spread.

We also perform qualitative analysis with visualization, plotting synthetic flight paths alongside real ones to assess whether shapes and patterns appear realistic. Such visual inspection, common in the evaluation of generative models \citep{murphy_probabilistic_2023}, can reveal implausible paths or insufficient variation that aggregate metrics may obscure.

%% file: sections/results.tex
\section{Results and Discussion}
%The following section contains the results from the transfer learning experiments.

\subsection{Transfer learning: rationale and setup}
We investigate whether knowledge learned at a data-rich airport transfers to a data-scarce airport, thereby reducing target-domain data requirements for realistic trajectory generation. This experiment follows the transfer-learning paradigm—pretraining on a source domain and fine-tuning on a target domain—to test cross-airport generalization of our generative models in the aviation setting (\citealp{zhu_difftraj_2023}). The objective is to quantify how much pretraining improves similarity to the target distribution when only limited target data are available. Models are first \emph{pretrained} on the Zürich (LSZH) landing dataset using the same configuration as in the baselines; they are then \emph{fine-tuned} on Dublin (EIDW) with data fractions $s\in\{0,5,20,50,100\}\%$ of the EIDW training split (fixed train/validation/test partitions and identical conditioning across runs). As runway labels are unavailable for Dublin, the runway condition is replaced by a single airport token (``EIDW''). For each split $s$, we train for 100 epochs (batch size 32), generate $N{=}100$ condition-matched samples, and evaluate against the held-out EIDW test set. % using the similarity metrics defined in Sec.~\ref{cha:ResearchAndResults:Eval}.

For each model family we compare a \emph{pretrained} model fine-tuned at $s\in\{0,5,20,50,100\}\%$ \emph{against a baseline trained from scratch on 100\% of EIDW}. This design probes how close a pretrained, low-data model can get to a strong full-data baseline at the target airport, thereby providing complementary evidence of transferability beyond local paired comparisons. Quantitative evaluation reports energy distance (e-distance), Maximum Mean Discrepancy (MMD), and Dynamic Time Warping (DTW), capturing distributional alignment and lateral shape agreement under consistent conditions. Metrics are computed on unsmoothed data; a secondary EWMA smoothing pass is used only for visualization in the longitude/latitude setting.

To support cross-airport generalization and avoid geographic leakage, the primary experiments operate in the \emph{track/groundspeed} parameterization (with altitude and $\Delta t$), reconstructing $(\mathrm{lat},\mathrm{lon})$ only for evaluation. This representation is explicitly \emph{airport-agnostic}: it encodes kinematics rather than absolute geography, requires no choice of map projection or local tangent plane, and is invariant to rigid transforms (translation/rotation) of the spatial frame. As such, it avoids geodesy/projection artifacts that arise with longitude/latitude when comparing distant airports and prevents leakage of location-specific geometry into the model \citep{snyder_map_1987,krauth_deep_2023}. The conditioning scheme is likewise agnostic to runway enumeration at the target: a single ``EIDW'' token replaces runway identity, while LSZH retains runway labels during pretraining.
For each target-data split we generate $N{=}100$ condition-matched samples and report mean $\pm$ standard deviation for e-distance, MMD, and DTW (lower is better). In the tables below, \textbf{bold} denotes a \emph{significant} improvement over the full-data Dublin baseline (Welch’s $t$-test, $p<0.05$). Metrics are computed on unsmoothed data; trajectory and t-SNE visualizations use the same plotting settings as in the thesis.

\subsection*{Diffusion (DM)}
\label{subsec:results:dm}

\begin{figure*}[!th]
    \centering
    \captionsetup[subfigure]{justification=centering}
    \begin{subfigure}[t]{0.24\textwidth}
        \includegraphics[width=\linewidth,trim={110 110 110 110},clip]{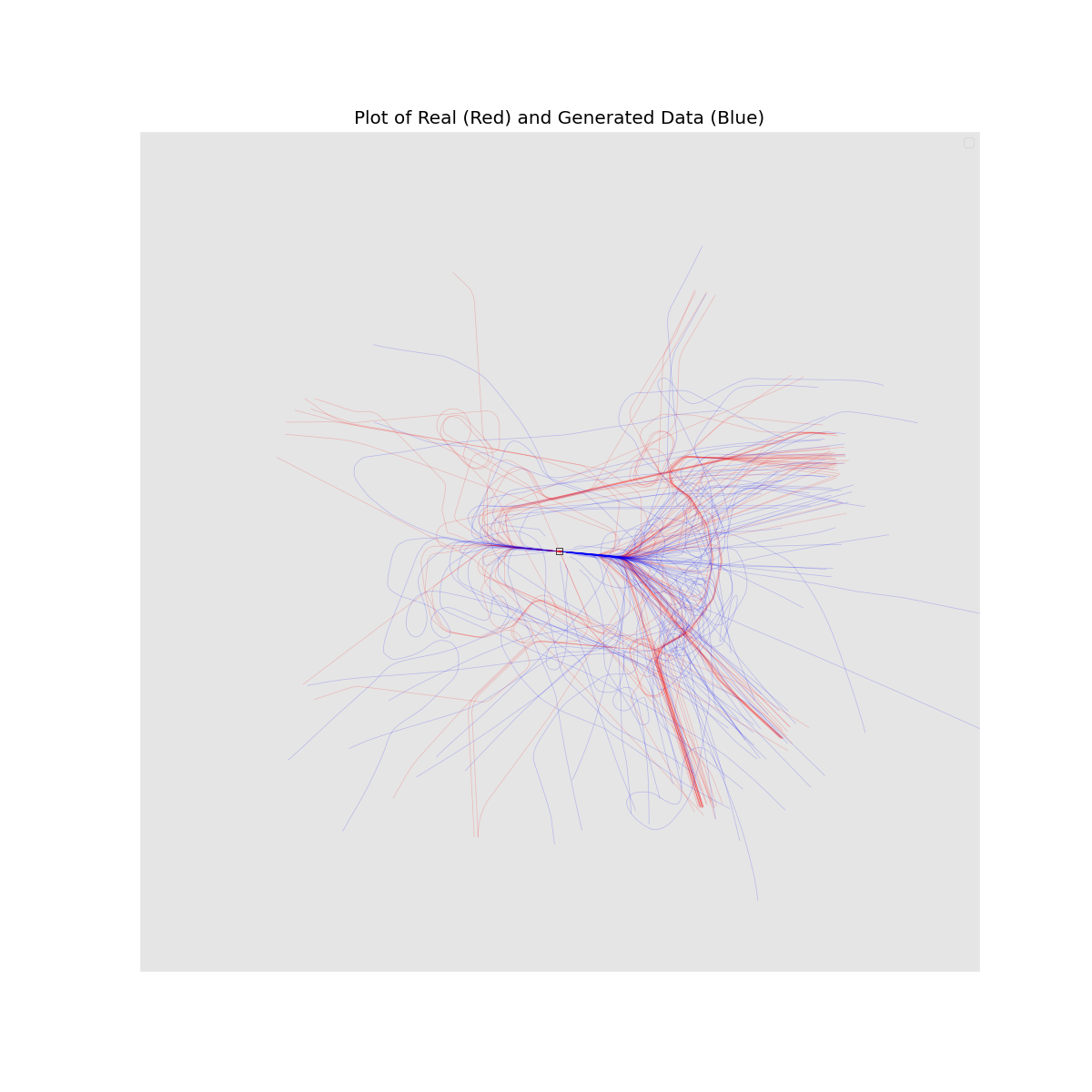}
        \caption{Baseline}
    \end{subfigure}\hfill
    \begin{subfigure}[t]{0.24\textwidth}
        \includegraphics[width=\linewidth,trim={130 140 130 160},clip]{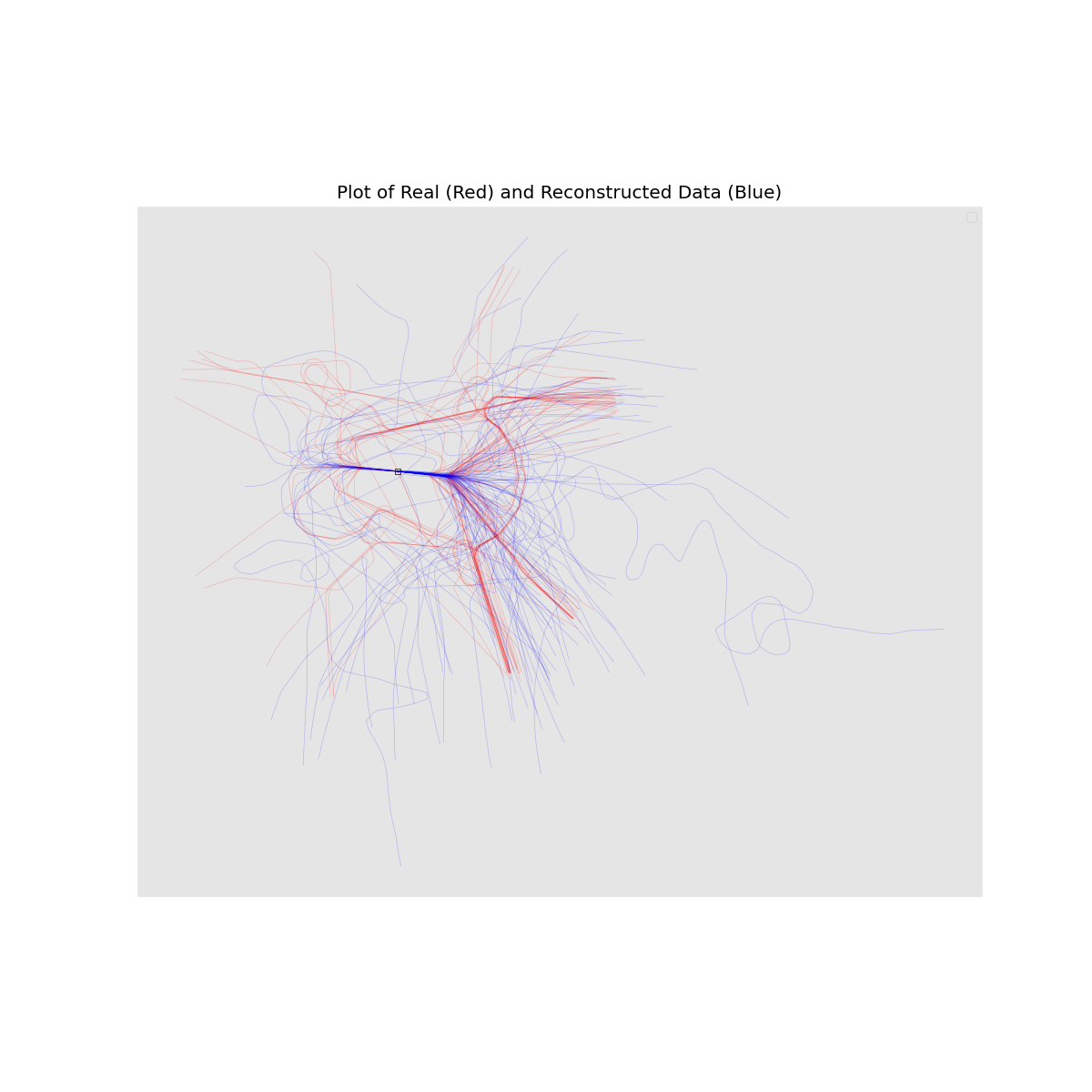}
        \caption{5\% split}
    \end{subfigure}\hfill
    \begin{subfigure}[t]{0.24\textwidth}
        \includegraphics[width=\linewidth,trim={110 110 110 110},clip]{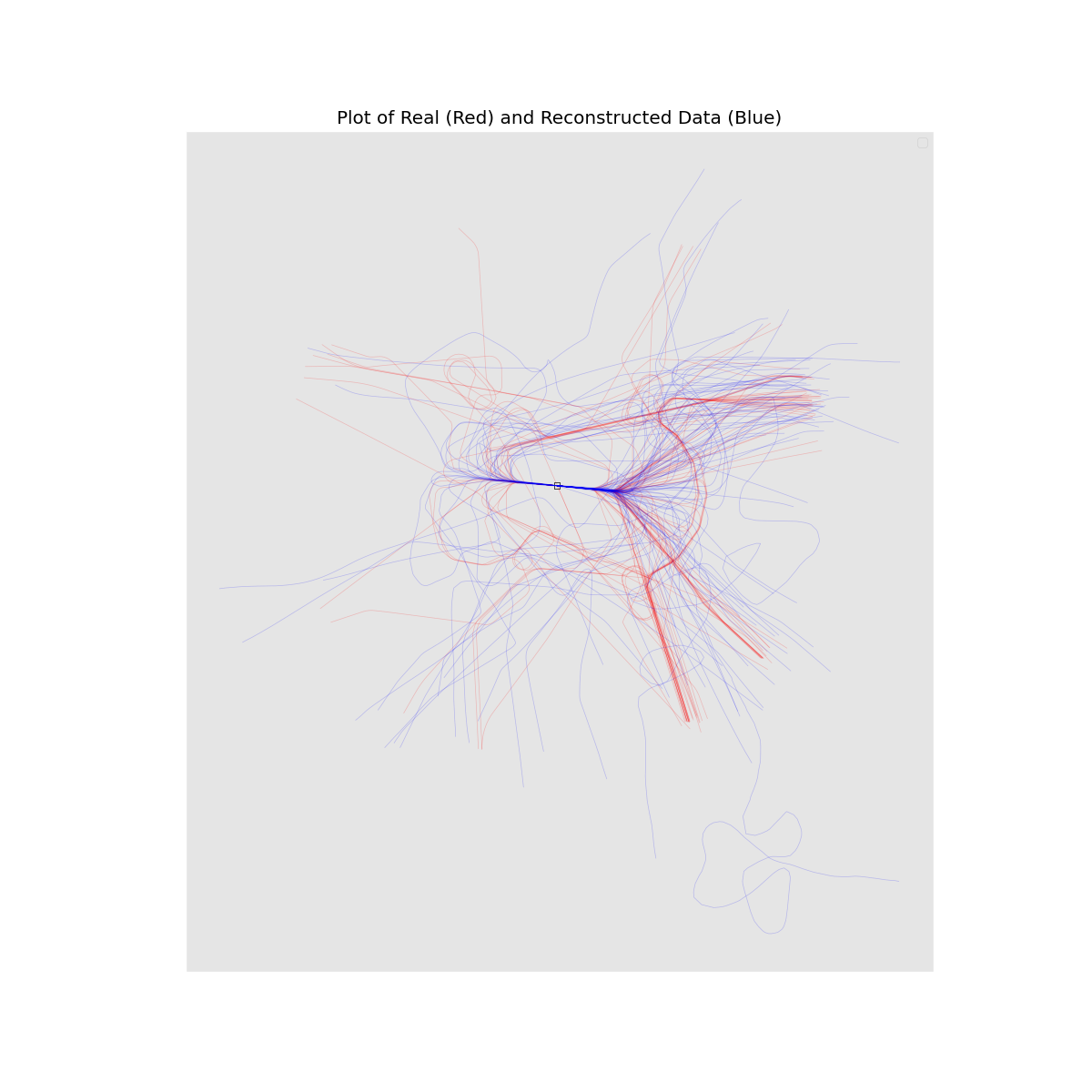}
        \caption{20\% split}
    \end{subfigure}\hfill
    \begin{subfigure}[t]{0.24\textwidth}
        \includegraphics[width=\linewidth,trim={110 110 110 110},clip]{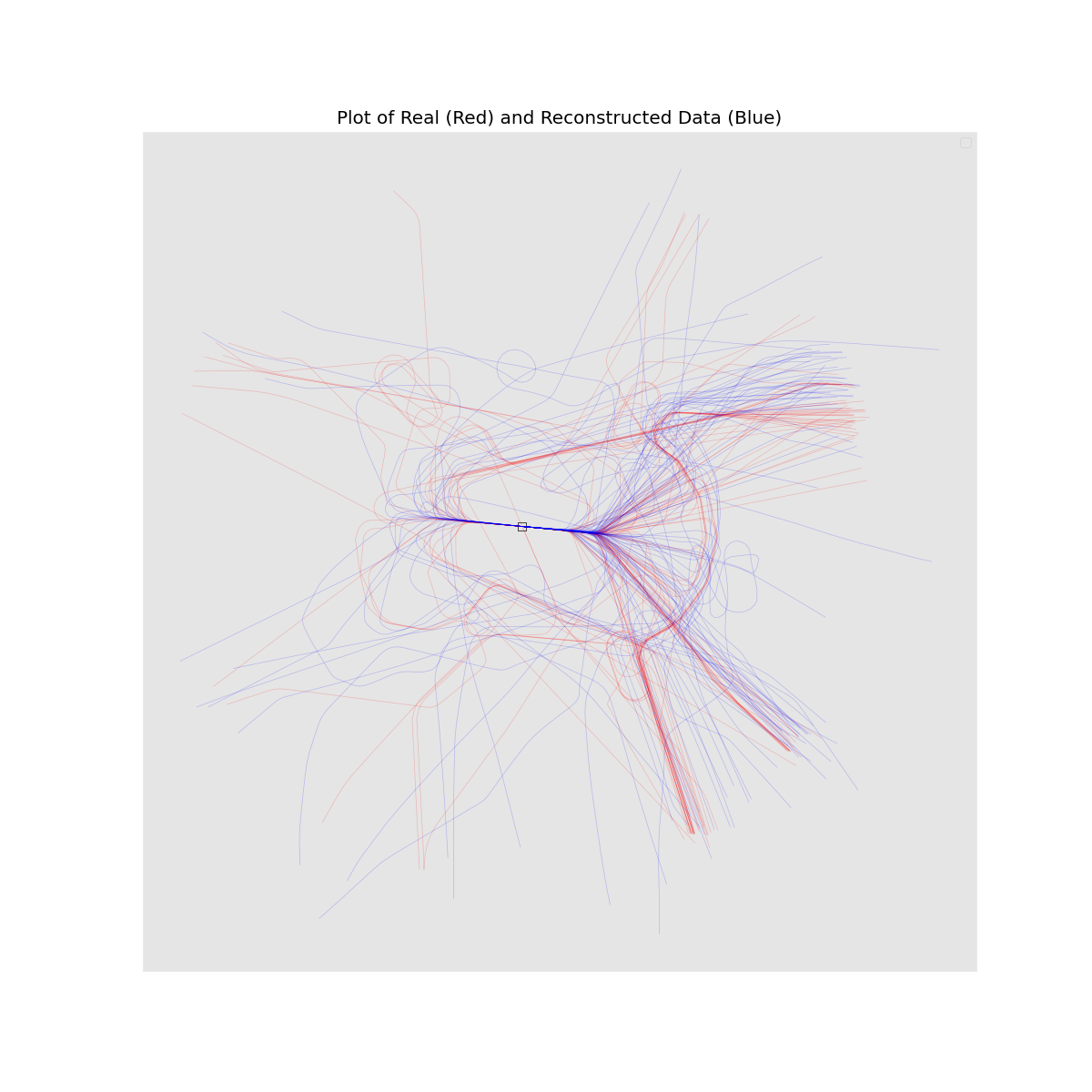}
        \caption{100\% split}
    \end{subfigure}
    \caption{DM: real (red) vs.\ generated (blue) trajectory overlays across Dublin data fractions.}
    \label{fig:dm_recons}
\end{figure*}

\begin{figure*}[!th]
    \centering
    \captionsetup[subfigure]{justification=centering}
    \begin{subfigure}[t]{0.25\textwidth}
        \includegraphics[width=\linewidth]{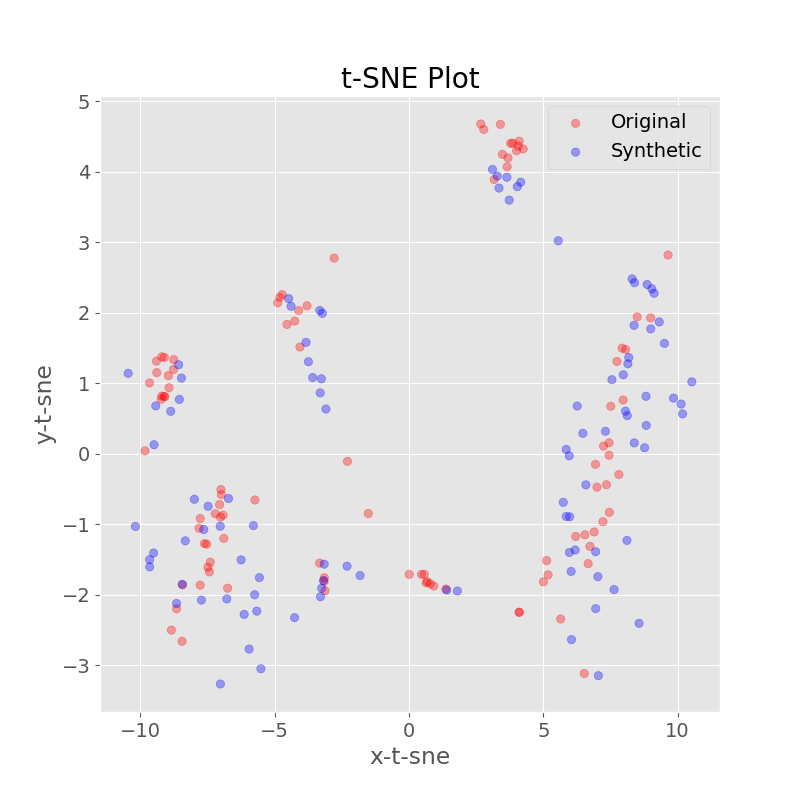}
        \caption{Baseline}
    \end{subfigure}\hfill
    \begin{subfigure}[t]{0.25\textwidth}
        \includegraphics[width=\linewidth]{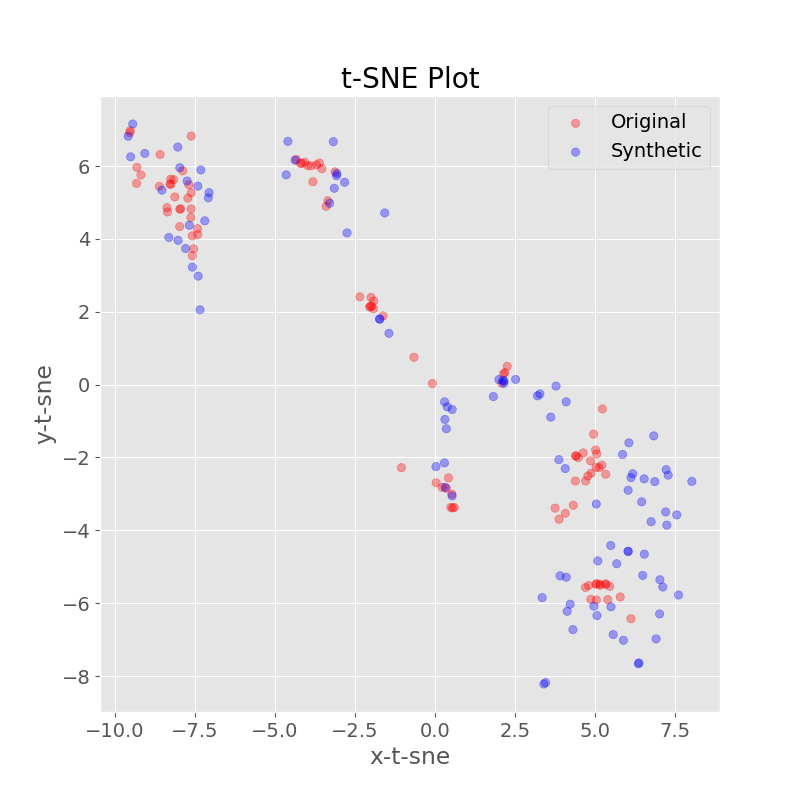}
        \caption{5\% split}
    \end{subfigure}\hfill
    \begin{subfigure}[t]{0.25\textwidth}
        \includegraphics[width=\linewidth]{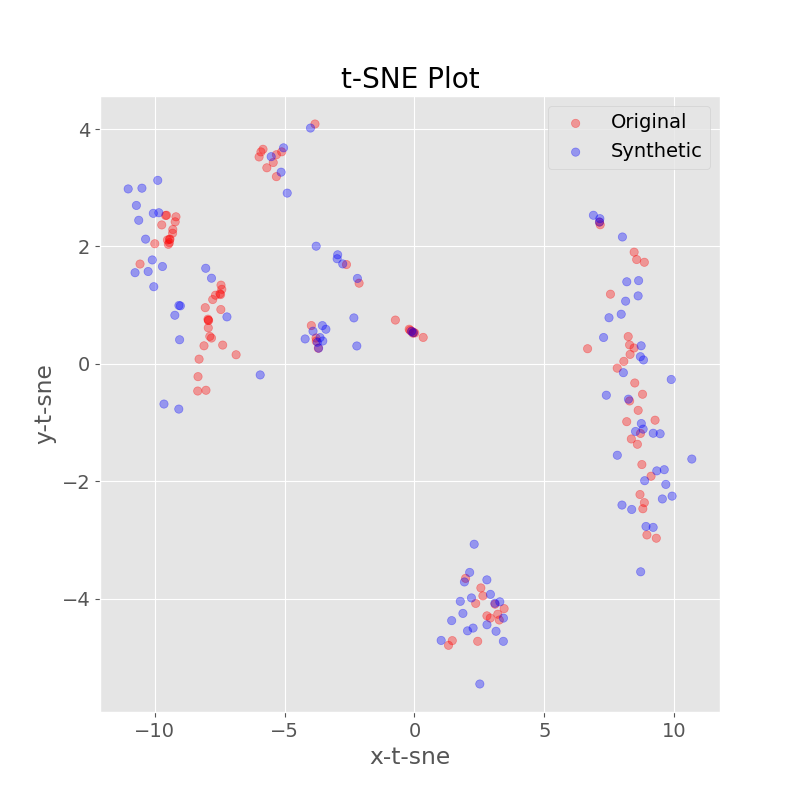}
        \caption{20\% split}
    \end{subfigure}\hfill
    \begin{subfigure}[t]{0.25\textwidth}
        \includegraphics[width=\linewidth]{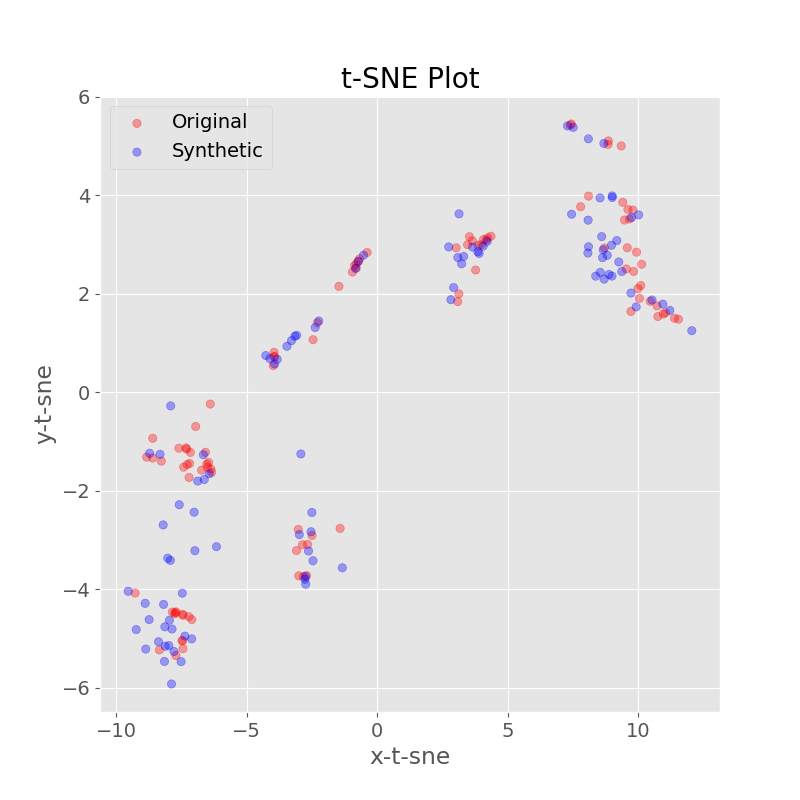}
        \caption{100\% split}
    \end{subfigure}
    \caption{DM: t-SNE overlays (real in red, generated in blue) across Dublin data fractions.}
    \label{fig:dm_tsne}
\end{figure*}

\autoref{tab:experiment:tl_track:dm} summarizes DM transfer performance. Zero-shot transfer ($0\%$) performs markedly worse than the Dublin baseline across all metrics, confirming that purely source-trained lateral structure does not carry to a new terminal area without \emph{any} target supervision. With only $5\%$ of Dublin labels, the pretrained DM already \emph{surpasses} the baseline on DTW (\textbf{23.21} vs.\ 29.10) and reduces MMD while keeping e-distance within a narrow margin. At $20\%$, the DM essentially closes the e-distance gap and achieves a further DTW reduction (\textbf{22.94}), indicating that fine-tuning aligns both global distributional fit and local path geometry. As the data budget grows to $50$–$100\%$, DTW falls steeply (\textbf{18.43} and \textbf{15.88}), while e-distance remains comparable to baseline. The small MMD uptick at larger splits is consistent with finite-sample variability with $N{=}100$ generations and does not appear in the visual diagnostics.

Qualitative evidence in \autoref{fig:dm_recons} shows that $5$–$20\%$ models already reproduce the dominant, densely flown corridors on the eastern side, with fewer artifacts at $20\%$ than at $5\%$. Residual errors concentrate in infrequent approach sectors (e.g., south–west entries), a pattern that persists, albeit attenuated, even at $100\%$. The t-SNE overlays in \autoref{fig:dm_tsne} echo this: overlap between synthetic and real clusters tightens monotonically with more Dublin data, while sparsely populated regions remain hardest to match. These findings are consistent with the thesis-wide observation that DM exhibits the strongest \emph{sample efficiency} among the tested architectures.

\begin{table}[h!]
\centering
\caption{DM transfer results on Dublin (track/groundspeed). Lower is better.}
\label{tab:experiment:tl_track:dm}
\begin{tabular}{lccc}
\toprule
\textbf{Split} & \textbf{e-distance} & \textbf{MMD} & \textbf{DTW} \\
\midrule
Baseline & 0.662 $\pm$ 0.170 & 0.113 $\pm$ 0.169 & 29.099 $\pm$ 0.106 \\
\midrule
0.00     & 1.228 $\pm$ 0.445 & 0.746 $\pm$ 0.207 & 118.222 $\pm$ 0.352 \\
0.05     & 0.693 $\pm$ 0.183 & 0.097 $\pm$ 0.139 & \textbf{23.211 $\pm$ 0.084} \\
0.20     & 0.681 $\pm$ 0.178 & 0.084 $\pm$ 0.135 & \textbf{22.941 $\pm$ 0.081} \\
0.50     & 0.679 $\pm$ 0.175 & 0.101 $\pm$ 0.131 & \textbf{18.428 $\pm$ 0.067} \\
1.00     & 0.668 $\pm$ 0.171 & 0.134 $\pm$ 0.142 & \textbf{15.876 $\pm$ 0.060} \\
\bottomrule
\end{tabular}
\end{table}

%\vspace{8pt}
\subsection*{Flow Matching (FM)}
\label{subsec:results:fm}
FM exhibits smaller and less consistent transfer gains than DM (\autoref{tab:experiment:tl_track:fm}). The zero-shot model underperforms the Dublin baseline on all metrics, though it degrades far less than DM at $0\%$, indicating a modest degree of cross-airport portability in the learned flow. With $5$–$20\%$ target data, e-distance and MMD trend downward relative to $0\%$, but remain broadly comparable to baseline and present higher variability. The $50\%$ split achieves a \emph{large} DTW improvement (\textbf{21.61}), yet coincides with a spike in MMD, suggesting a sharper geometric fit along a subset of approach corridors at the expense of distributional coverage. At $100\%$, e-distance and MMD align with baseline while DTW remains significantly lower (\textbf{24.98}), indicating improved local path geometry but limited overall distributional gains.

\begin{figure*}[!th]
    \centering
    \captionsetup[subfigure]{justification=centering}
    \begin{subfigure}[t]{0.24\textwidth}
        \includegraphics[width=\linewidth,trim={110 200 110 110},clip]{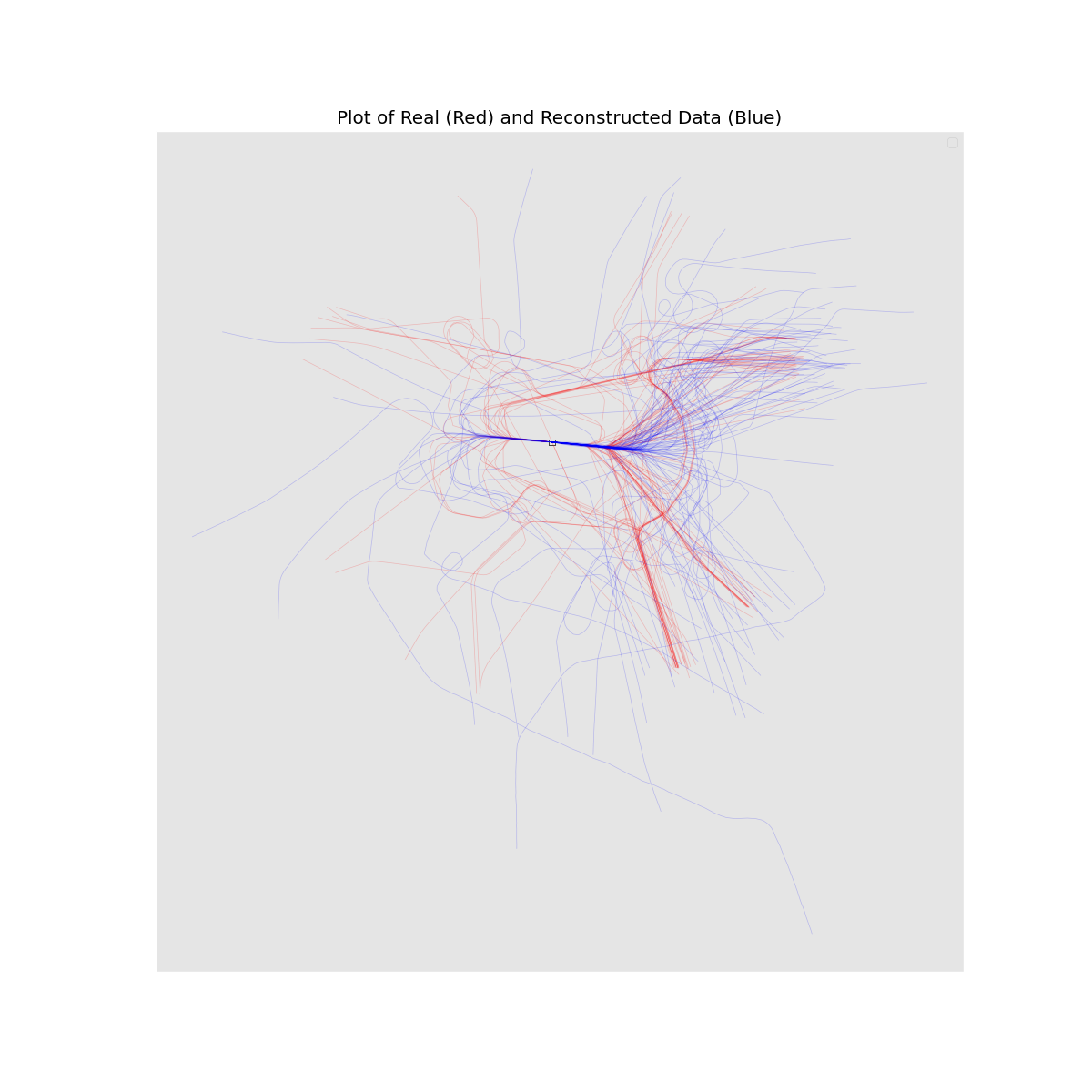}
        \caption{Baseline}
    \end{subfigure}\hfill
    \begin{subfigure}[t]{0.24\textwidth}
        \includegraphics[width=\linewidth,trim={110 110 110 240},clip]{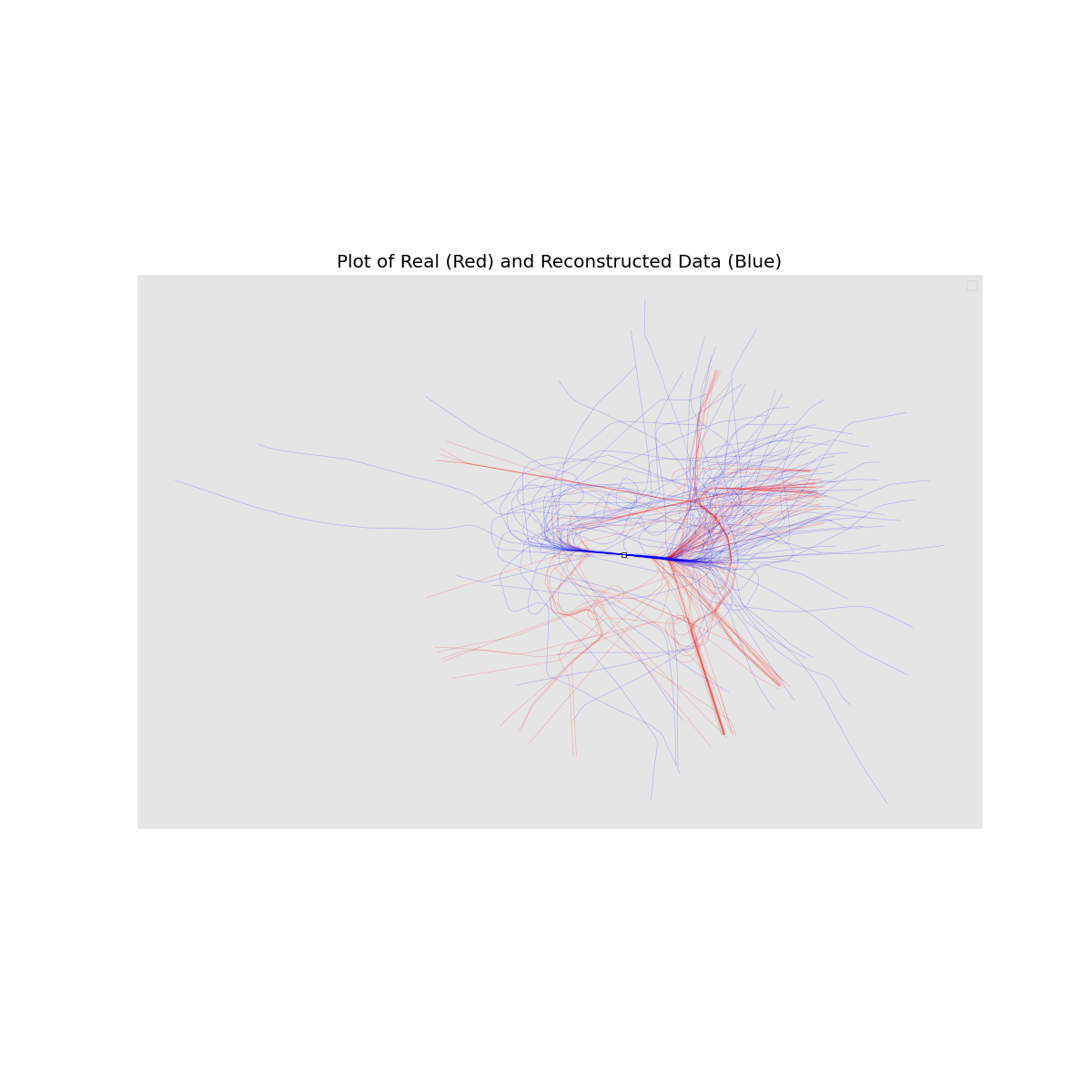}
        \caption{5\% split}
    \end{subfigure}\hfill
    \begin{subfigure}[t]{0.24\textwidth}
        \includegraphics[width=\linewidth,trim={110 110 110 200},clip]{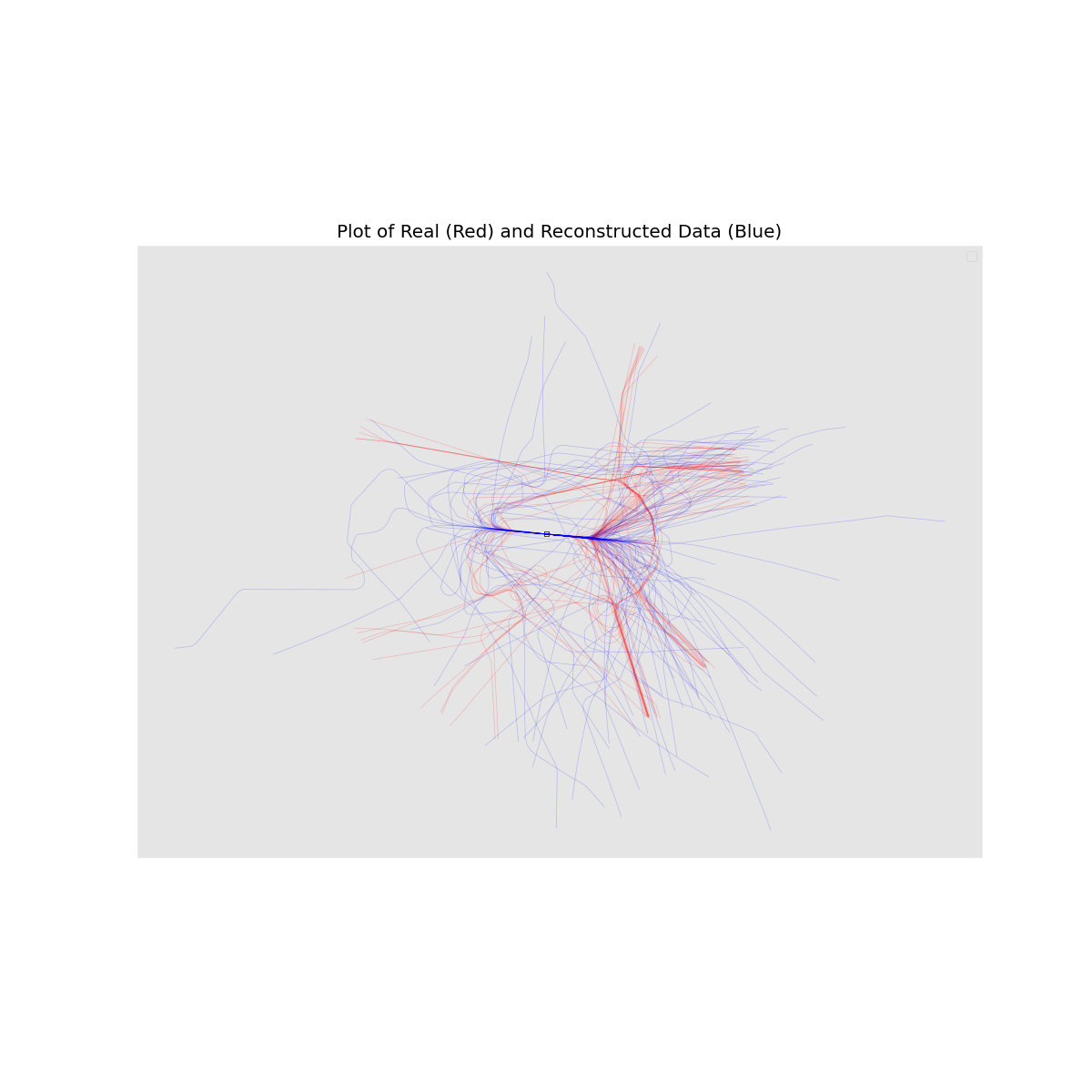}
        \caption{20\% split}
    \end{subfigure}\hfill
    \begin{subfigure}[t]{0.24\textwidth}
        \includegraphics[width=\linewidth,trim={110 110 110 240},clip]{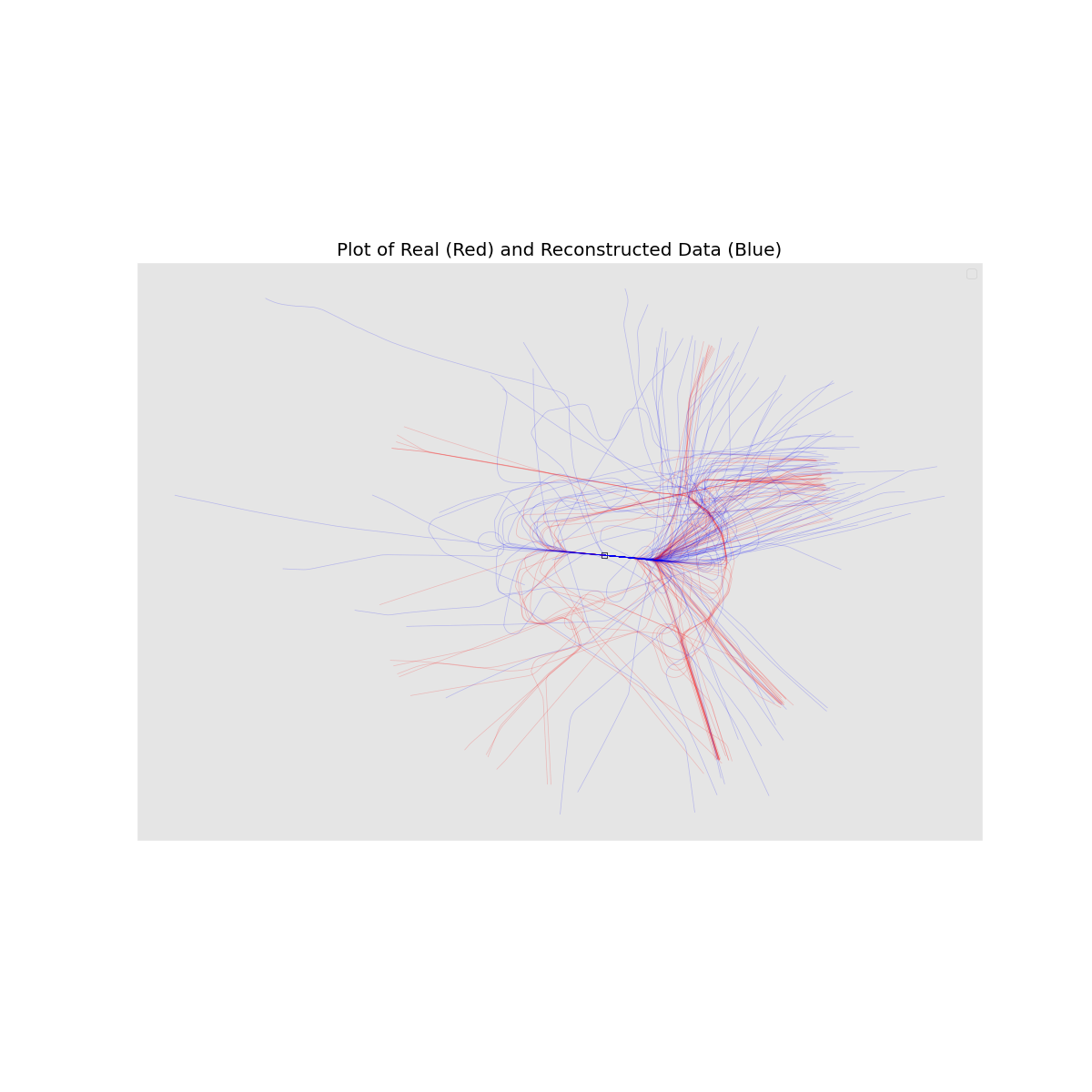}
        \caption{100\% split}
    \end{subfigure}
    \caption{FM: real (red) vs.\ generated (blue) trajectory overlays across Dublin data fractions.}
    \label{fig:fm_recons}
\end{figure*}

\begin{figure*}[!th]
    \centering
    \captionsetup[subfigure]{justification=centering}
    \begin{subfigure}[t]{0.25\textwidth}
        \includegraphics[width=\linewidth]{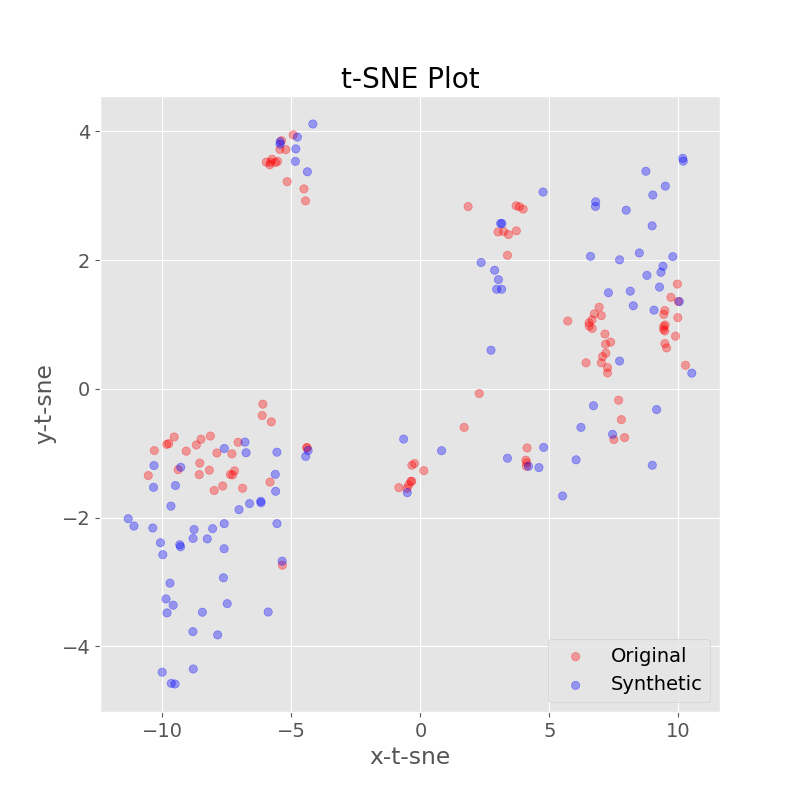}
        \caption{Baseline}
    \end{subfigure}\hfill
    \begin{subfigure}[t]{0.25\textwidth}
        \includegraphics[width=\linewidth]{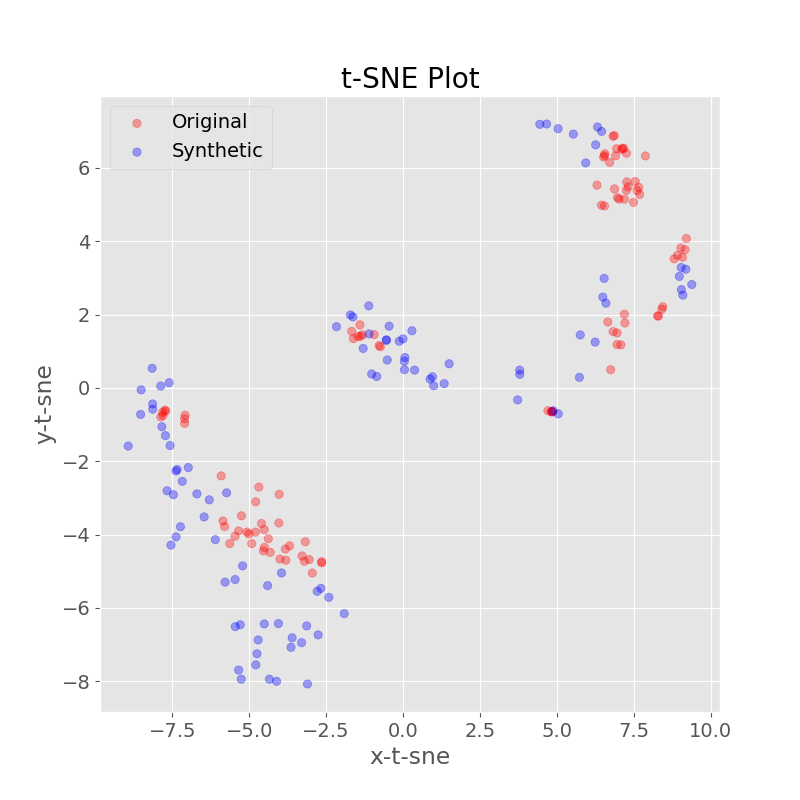}
        \caption{5\% split}
    \end{subfigure}\hfill
    \begin{subfigure}[t]{0.25\textwidth}
        \includegraphics[width=\linewidth]{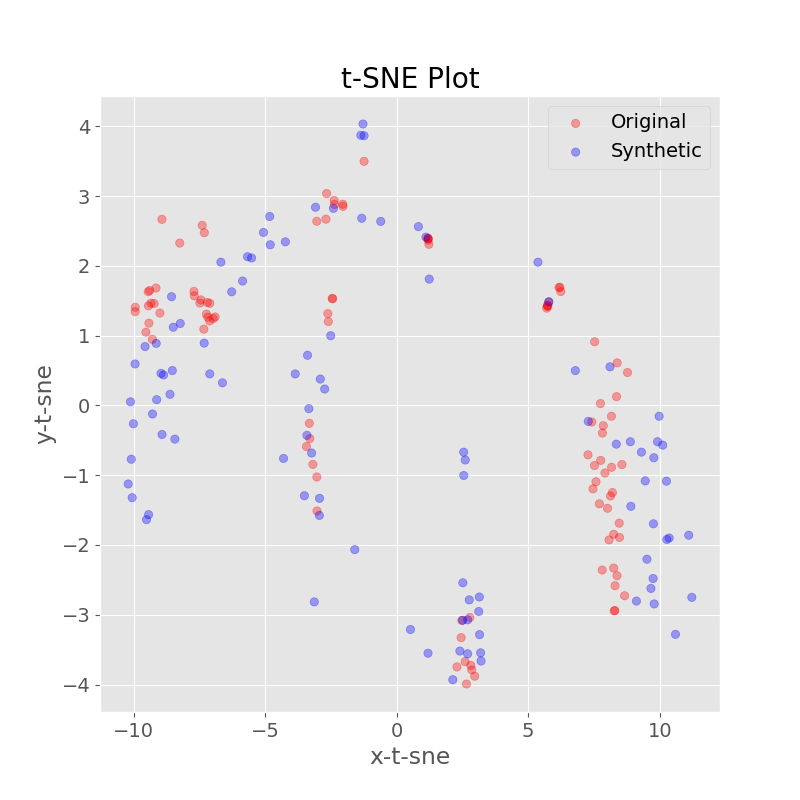}
        \caption{20\% split}
    \end{subfigure}\hfill
    \begin{subfigure}[t]{0.25\textwidth}
        \includegraphics[width=\linewidth]{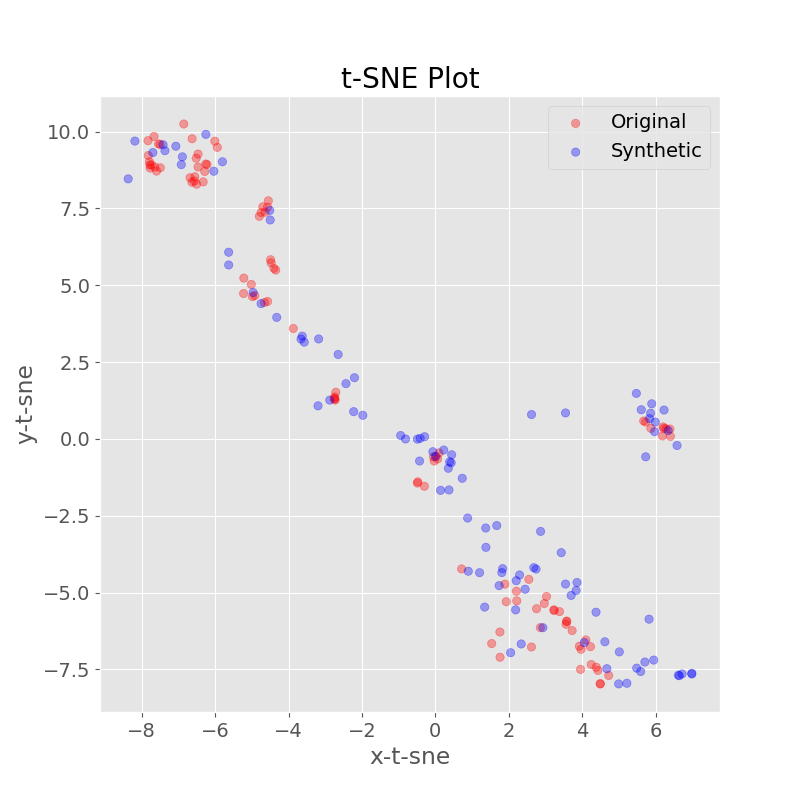}
        \caption{100\% split}
    \end{subfigure}
    \caption{FM: t-SNE overlays (real in red, generated in blue) across Dublin data fractions.}
    \label{fig:fm_tsne}
\end{figure*}

Trajectory overlays in \autoref{fig:fm_recons} show that FM tends to favor specific high-density sectors (notably the upper-right approach), producing more outliers on the left-hand side of the terminal area. The $0.20$ model spreads coverage more evenly but introduces noise, consistent with its low MMD yet modest DTW. The $1.00$ model reduces artifacts and yields the visually cleanest paths, while still underrepresenting rarer west-facing approaches. The t-SNE plots in \autoref{fig:fm_tsne} confirm the picture: partial cluster coverage at $0.05$–$0.20$, improved but incomplete overlap at $1.00$, and residual outliers. Overall, FM benefits from transfer, but its gains are smaller and more variable than DM’s, in line with the thesis baseline where FM lagged in raw performance.

\begin{table}[t!]
\centering
\caption{FM transfer results on Dublin (track/groundspeed). Lower is better.}
\label{tab:experiment:tl_track:fm}
\begin{tabular}{lccc}
\toprule
\textbf{Split} & \textbf{e-distance} & \textbf{MMD} & \textbf{DTW} \\
\midrule
Baseline & 0.694 $\pm$ 0.180 & 0.094 $\pm$ 0.153 & 26.554 $\pm$ 0.098 \\
\midrule
0.00     & 0.750 $\pm$ 0.216 & 0.368 $\pm$ 0.179 & 37.490 $\pm$ 0.125 \\
0.05     & 0.720 $\pm$ 0.190 & 0.115 $\pm$ 0.161 & 28.859 $\pm$ 0.100 \\
0.20     & 0.713 $\pm$ 0.187 & 0.071 $\pm$ 0.160 & 27.397 $\pm$ 0.094 \\
0.50     & 0.676 $\pm$ 0.177 & 0.262 $\pm$ 0.171 & \textbf{21.607 $\pm$ 0.081} \\
1.00     & 0.697 $\pm$ 0.186 & 0.081 $\pm$ 0.154 & \textbf{24.975 $\pm$ 0.092} \\
\bottomrule
\end{tabular}
\end{table}

%\vspace{8pt}
\subsection*{Latent Diffusion (LDM)}
\label{subsec:results:ldm}

\begin{figure*}[!th]
    \centering
    \captionsetup[subfigure]{justification=centering}
    \begin{subfigure}[t]{0.24\textwidth}
        \includegraphics[width=\linewidth,trim={110 110 110 150},clip]{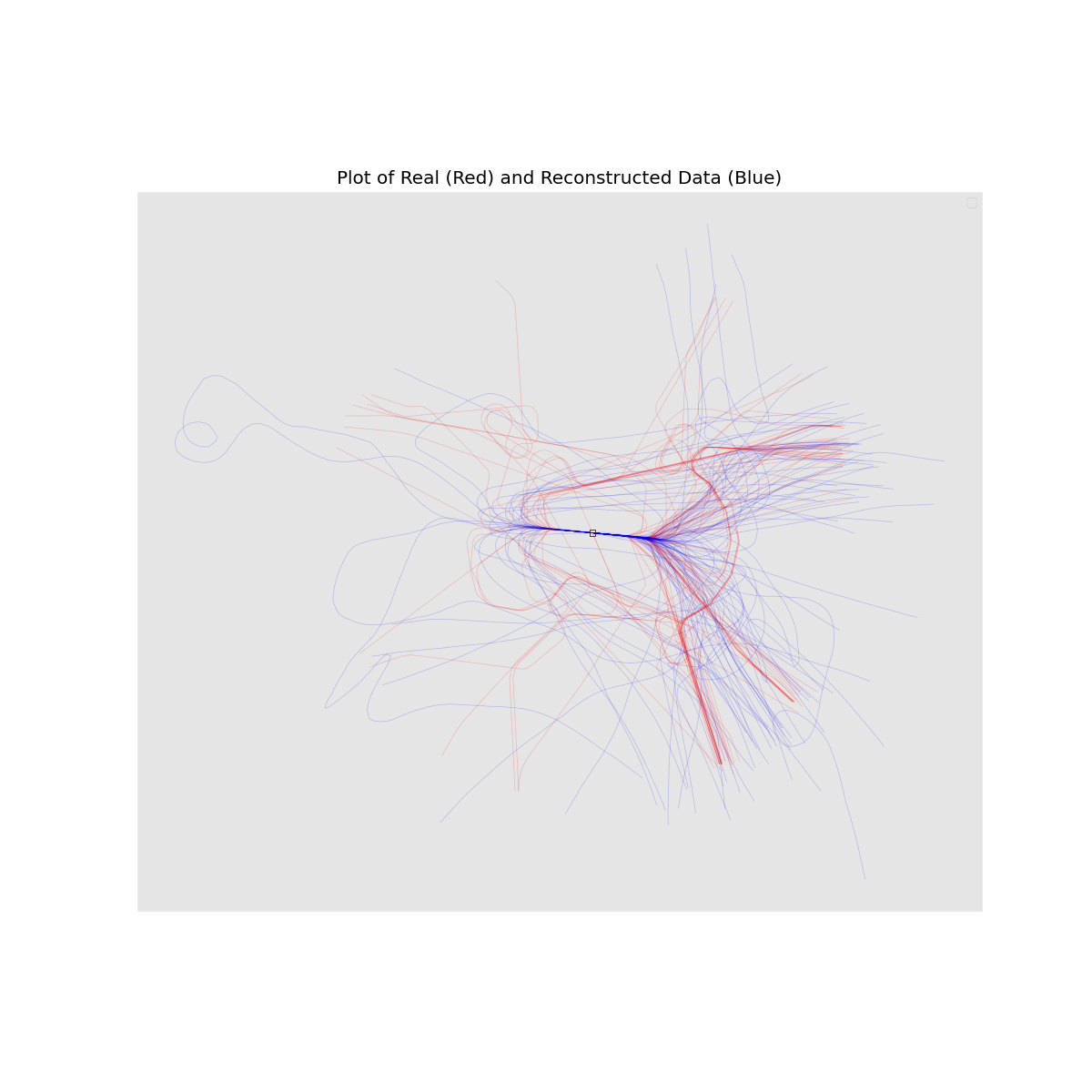}
        \caption{Baseline}
    \end{subfigure}\hfill
    \begin{subfigure}[t]{0.24\textwidth}
        \includegraphics[width=\linewidth,trim={110 110 110 190},clip]{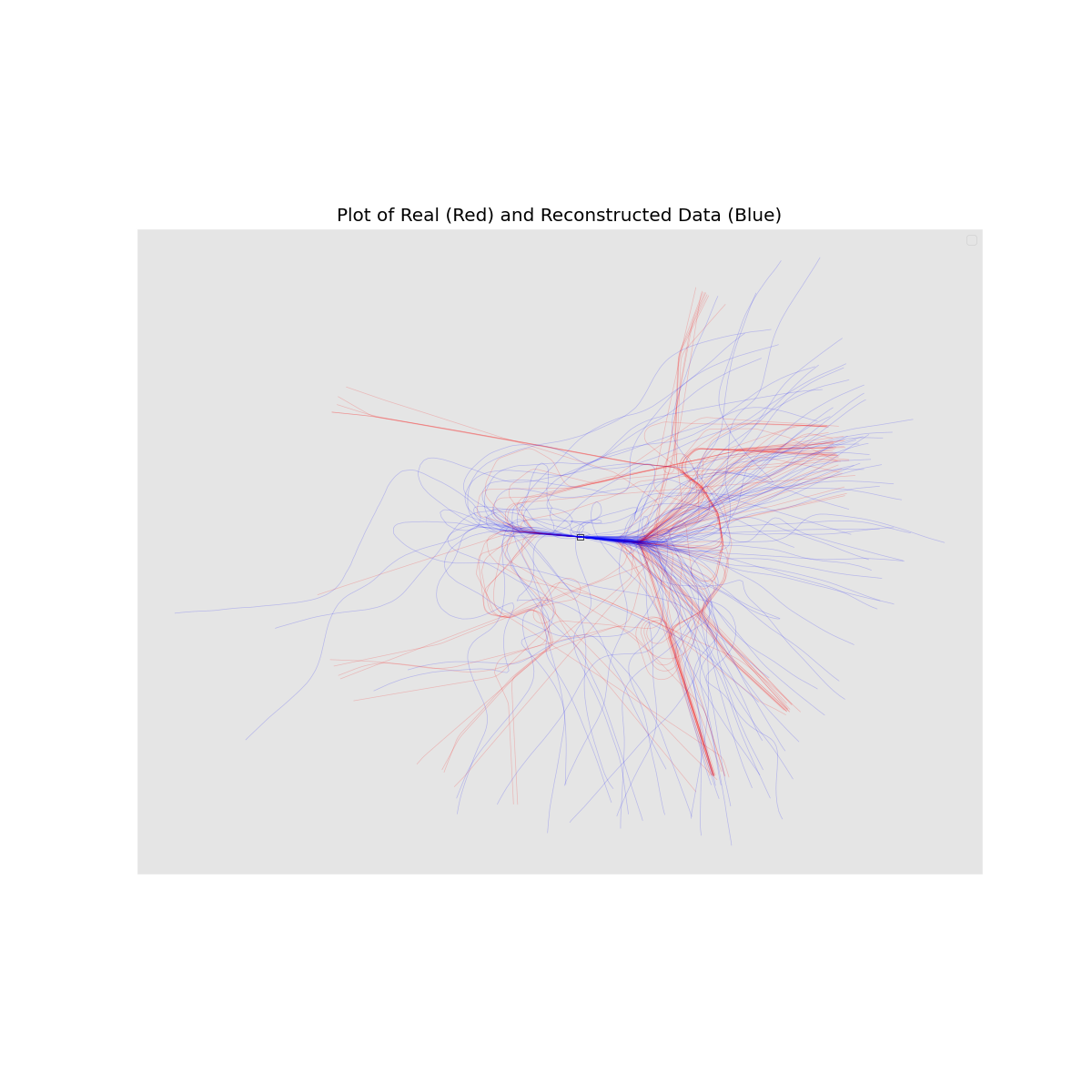}
        \caption{5\% split}
    \end{subfigure}\hfill
    \begin{subfigure}[t]{0.24\textwidth}
        \includegraphics[width=\linewidth,trim={110 110 110 160},clip]{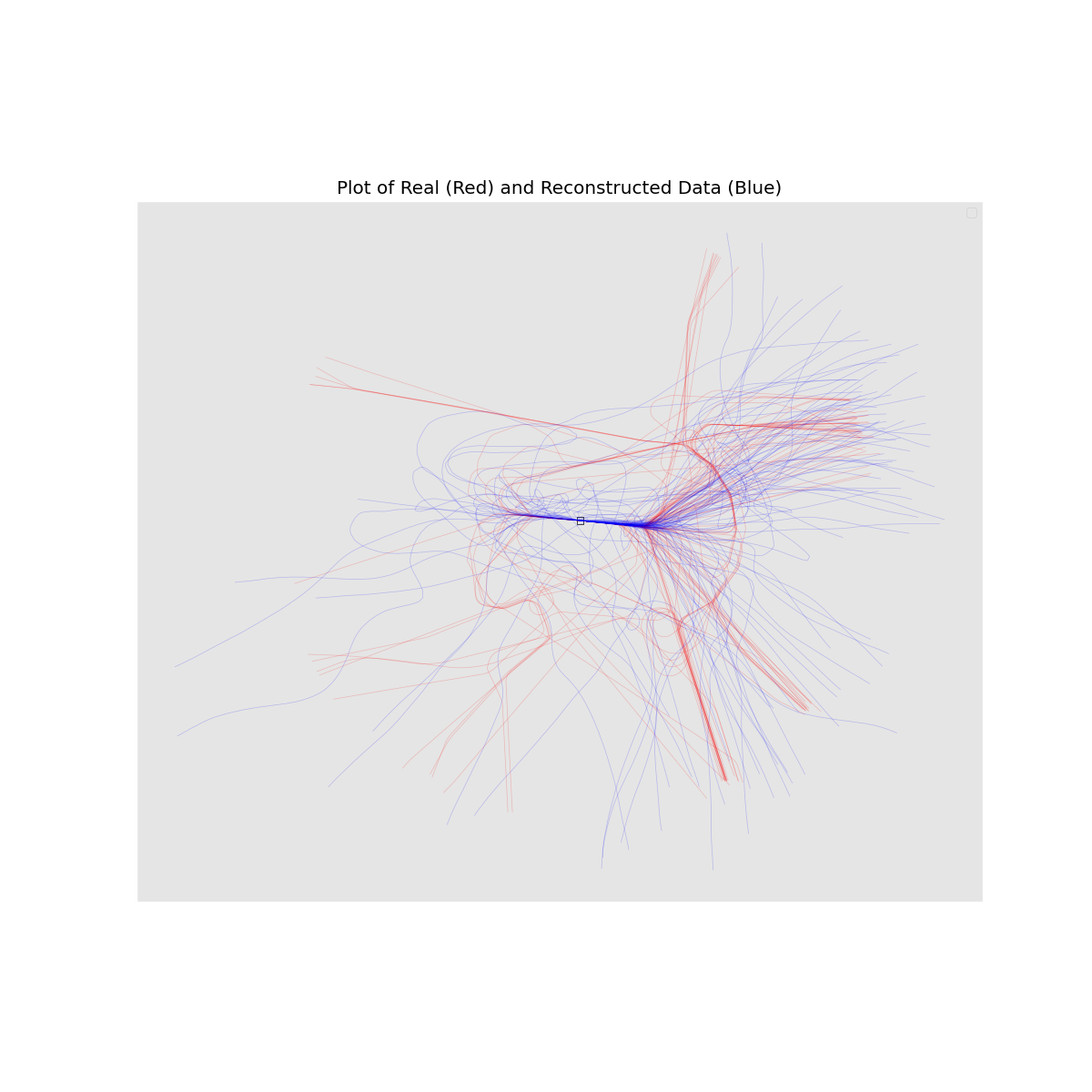}
        \caption{20\% split}
    \end{subfigure}\hfill
    \begin{subfigure}[t]{0.24\textwidth}
        \includegraphics[width=\linewidth,trim={110 110 110 110},clip]{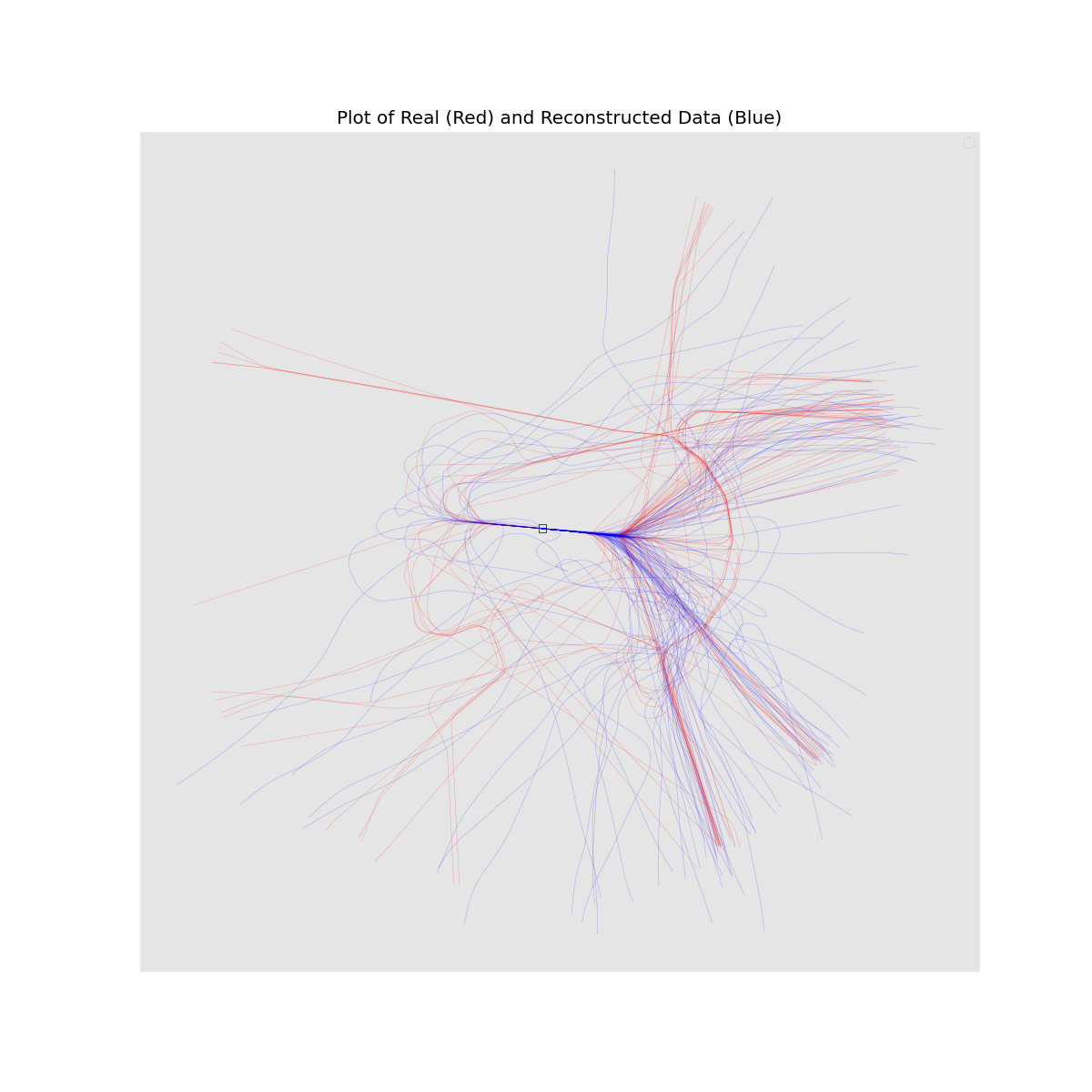}
        \caption{100\% split}
    \end{subfigure}
    \caption{LDM: real (red) vs.\ generated (blue) trajectory overlays across Dublin data fractions.}
    \label{fig:ldm_recons}
\end{figure*}

\begin{figure*}[!th]
    \centering
    \captionsetup[subfigure]{justification=centering}
    \begin{subfigure}[t]{0.25\textwidth}
        \includegraphics[width=\linewidth]{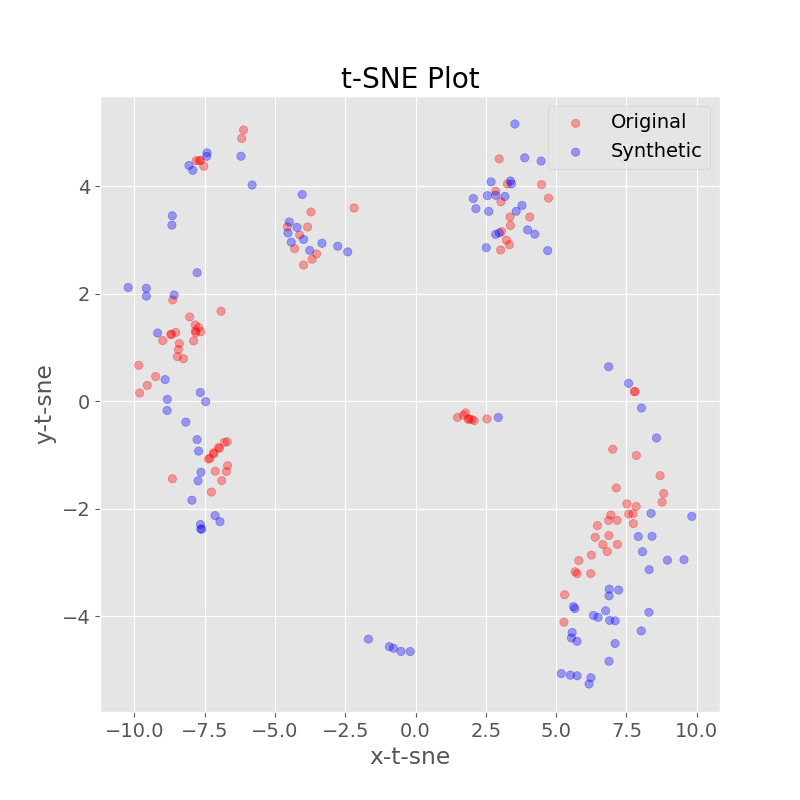}
        \caption{Baseline}
    \end{subfigure}\hfill
    \begin{subfigure}[t]{0.25\textwidth}
        \includegraphics[width=\linewidth]{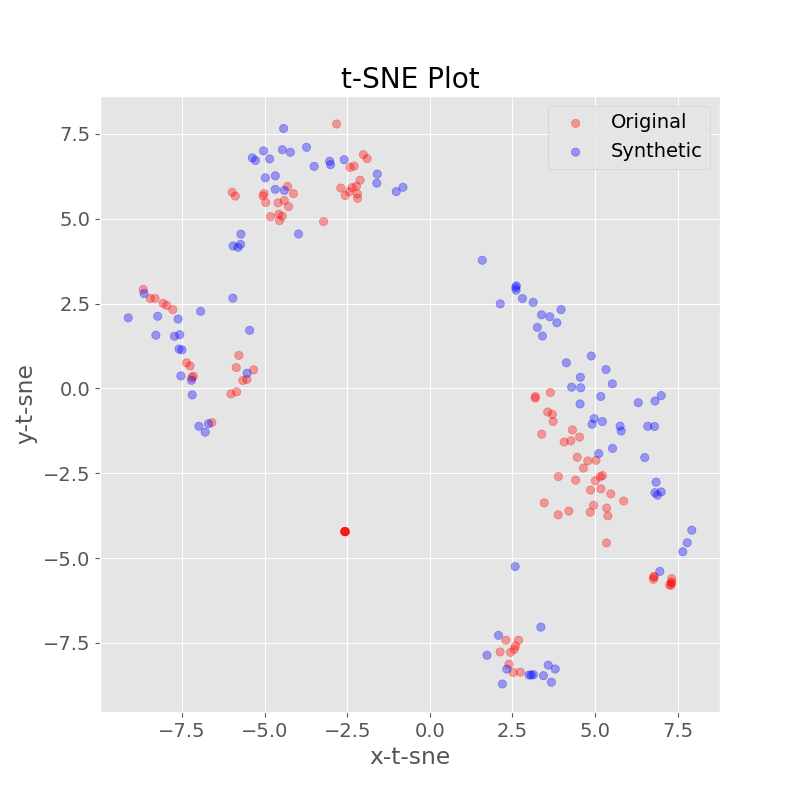}
        \caption{5\% split}
    \end{subfigure}\hfill
    \begin{subfigure}[t]{0.25\textwidth}
        \includegraphics[width=\linewidth]{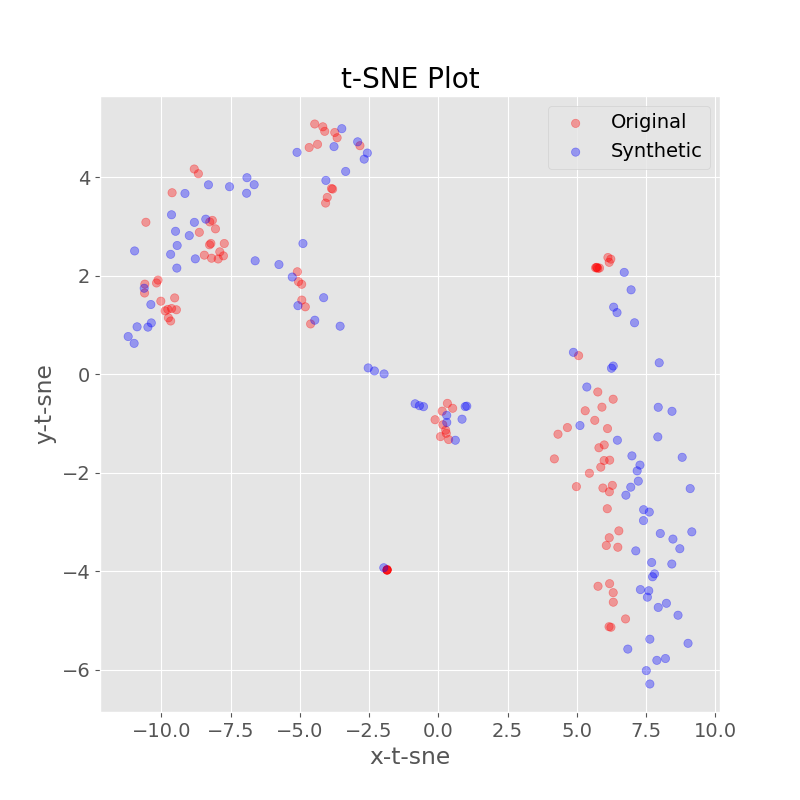}
        \caption{20\% split}
    \end{subfigure}\hfill
    \begin{subfigure}[t]{0.25\textwidth}
        \includegraphics[width=\linewidth]{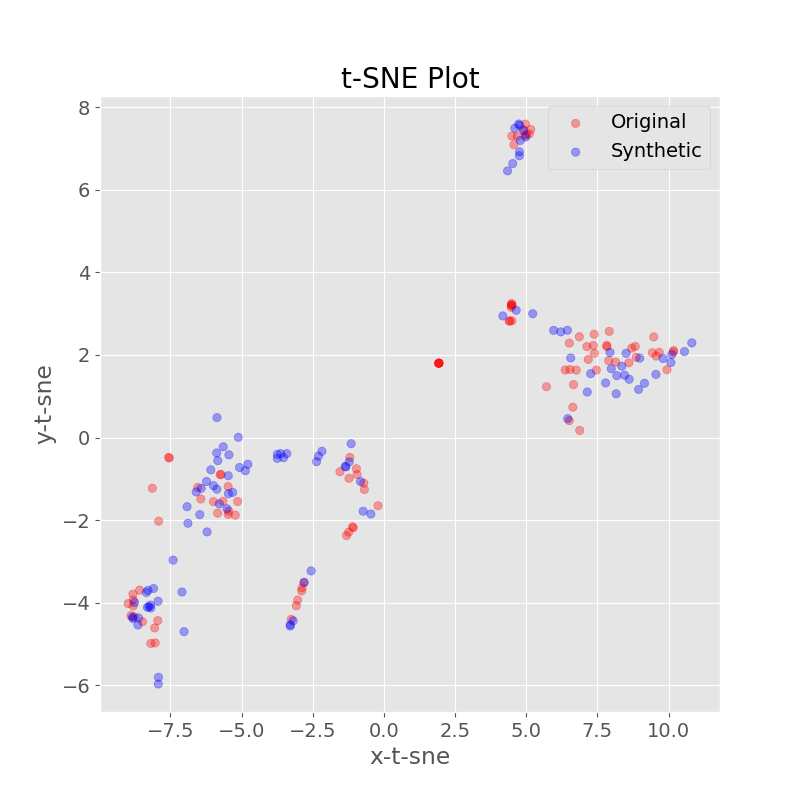}
        \caption{100\% split}
    \end{subfigure}
    \caption{LDM: t-SNE overlays (real in red, generated in blue) across Dublin data fractions.}
    \label{fig:ldm_tsne}
\end{figure*}

LDM shows the same qualitative pattern as DM but with smaller effect sizes (\autoref{tab:experiment:tl_track:ldm}). Without target data ($0\%$), performance lags the baseline, reflecting the need to adapt the latent generative mapping to local arrival structure. With $5\%$ data, LDM narrows the gap; at $20\%$ it reaches baseline-level e-distance/MMD and achieves a \emph{significant} DTW improvement (\textbf{21.40} vs.\ 22.78). Larger splits ($50$–$100\%$) further reduce DTW (\textbf{19.64}, \textbf{20.04}); MMD decreases up to $20\%$ then increases—consistent with higher sensitivity of kernel metrics in finite-$N$ settings.

Visually (\autoref{fig:ldm_recons}), fine-tuning reduces outliers and stabilizes path geometry, particularly on the high-density eastern approaches; west-facing trajectories remain comparatively underrepresented, consistent with their lower frequency in the data. The t-SNE overlays (\autoref{fig:ldm_tsne}) show steadily improving coverage of real clusters from $5\%$ to $100\%$, with residual gaps confined to sparse regions. LDM thus benefits from pretraining and target adaptation, delivering robust DTW gains at moderate data budgets, though it does not match DM’s sample efficiency.

\begin{table}[h!]
\centering
\caption{LDM transfer results on Dublin (track/groundspeed). Lower is better.}
\label{tab:experiment:tl_track:ldm}
\begin{tabular}{lccc}
\toprule
\textbf{Split} & \textbf{e-distance} & \textbf{MMD} & \textbf{DTW} \\
\midrule
Baseline & 0.683 $\pm$ 0.173 & 0.128 $\pm$ 0.126 & 22.779 $\pm$ 0.084 \\
\midrule
0.00     & 0.743 $\pm$ 0.227 & 0.201 $\pm$ 0.108 & 34.800 $\pm$ 0.122 \\
0.05     & 0.691 $\pm$ 0.179 & 0.162 $\pm$ 0.142 & 24.908 $\pm$ 0.088 \\
0.20     & 0.683 $\pm$ 0.178 & 0.133 $\pm$ 0.144 & \textbf{21.401 $\pm$ 0.076} \\
0.50     & 0.669 $\pm$ 0.174 & 0.188 $\pm$ 0.160 & \textbf{19.640 $\pm$ 0.074} \\
1.00     & 0.667 $\pm$ 0.172 & 0.234 $\pm$ 0.144 & \textbf{20.038 $\pm$ 0.078} \\
\bottomrule
\end{tabular}
\end{table}

%\vspace{8pt}
\subsection*{Latent Flow Matching (LFM)}
\label{subsec:results:lfm}
Among latent models, LFM starts from the weakest Dublin baseline but gains the most from transfer (\autoref{tab:experiment:tl_track:lfm}). With $5\%$ data, both e-distance and DTW significantly improve over the baseline, indicating that a small amount of target supervision suffices to tune the latent flow to local structure. At $20\%$ the improvements strengthen across \emph{all} metrics, including a pronounced MMD drop (from 0.168 to \textbf{0.081}), which suggests better coverage of the high-density regions. As the budget grows to $50$–$100\%$, DTW continues to decrease (\textbf{34.46} and \textbf{27.43}), and MMD reaches \textbf{0.060} at $100\%$. These results are consistent with the thesis insight that LFM benefits disproportionately from exposure to \emph{diverse} target examples.

\begin{figure*}[!th]
    \centering
    \captionsetup[subfigure]{justification=centering}
    \begin{subfigure}[t]{0.24\textwidth}
        \includegraphics[width=\linewidth,trim={110 110 110 160},clip]{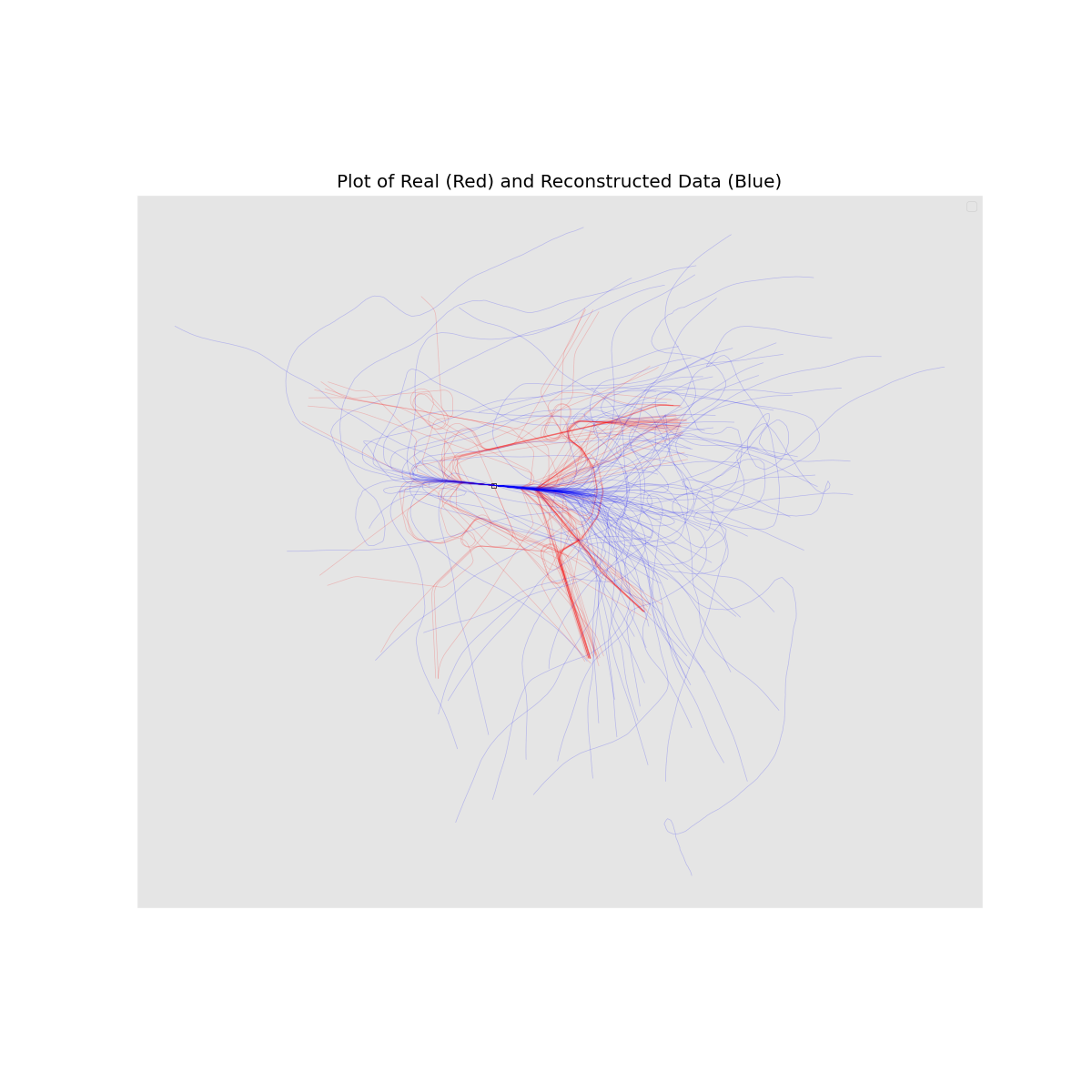}
        \caption{Baseline}
    \end{subfigure}\hfill
    \begin{subfigure}[t]{0.24\textwidth}
        \includegraphics[width=\linewidth,trim={110 110 110 110},clip]{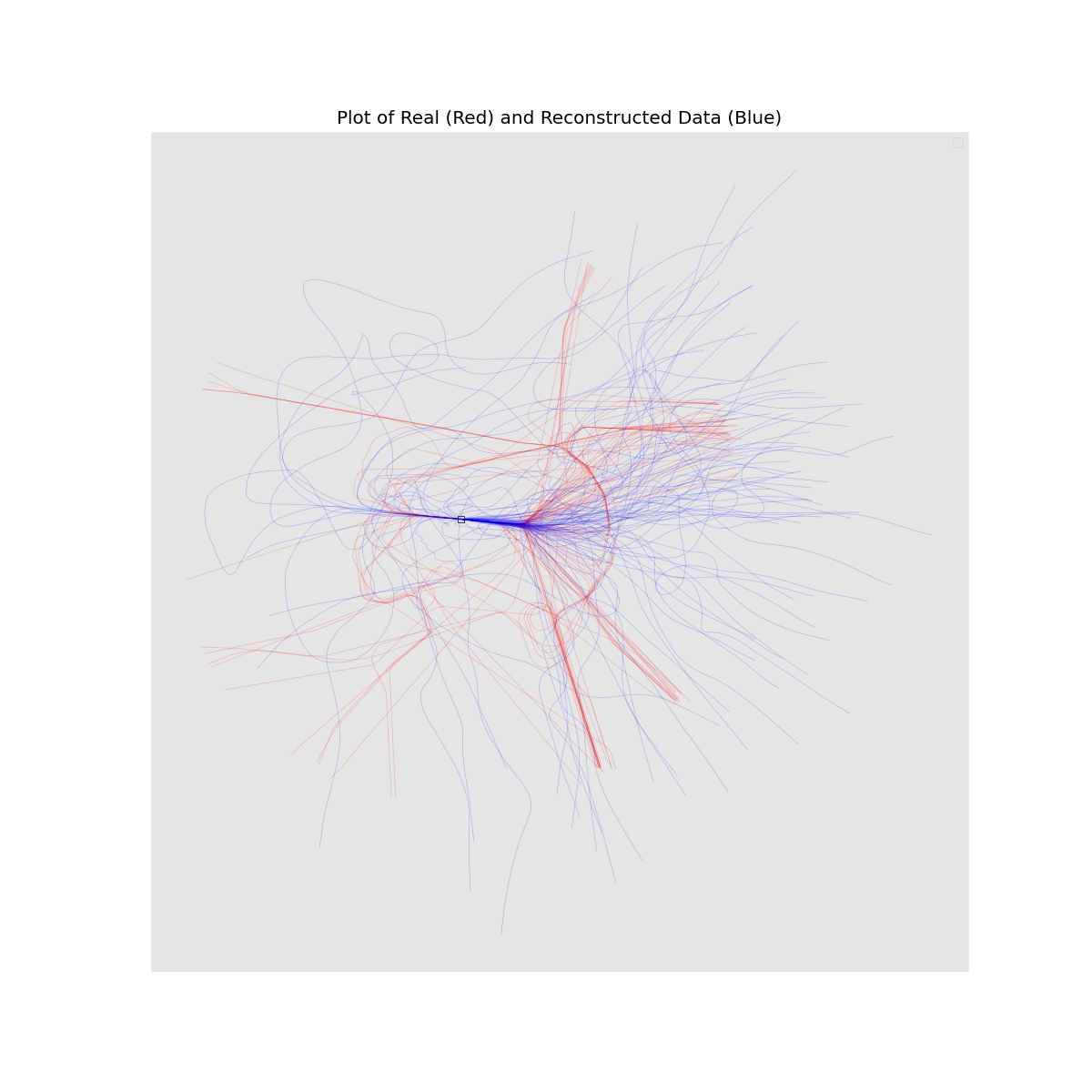}
        \caption{5\% split}
    \end{subfigure}\hfill
    \begin{subfigure}[t]{0.24\textwidth}
        \includegraphics[width=\linewidth,trim={110 110 110 110},clip]{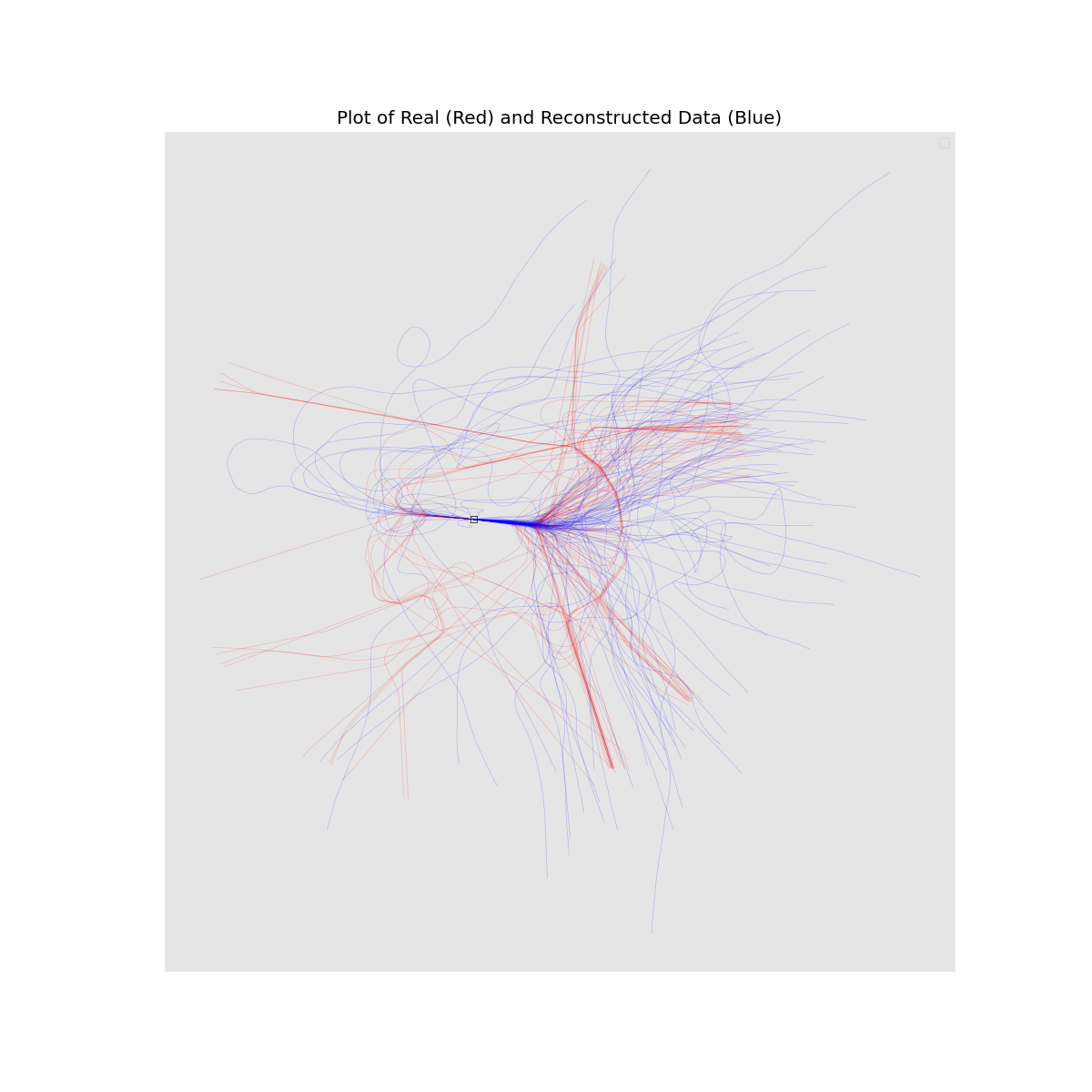}
        \caption{20\% split}
    \end{subfigure}\hfill
    \begin{subfigure}[t]{0.24\textwidth}
        \includegraphics[width=\linewidth,trim={110 110 110 110},clip]{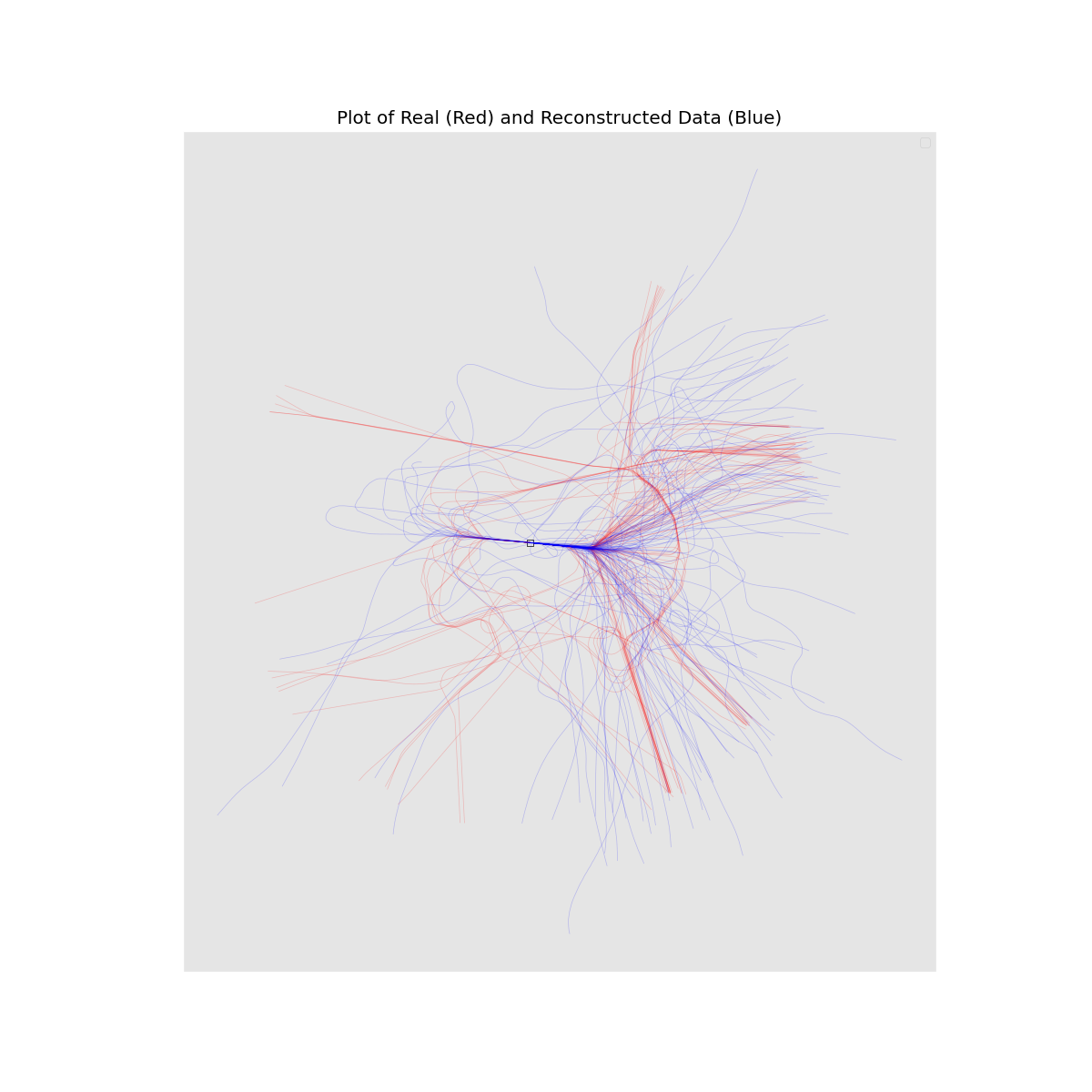}
        \caption{100\% split}
    \end{subfigure}
    \caption{LFM: real (red) vs.\ generated (blue) trajectory overlays across Dublin data fractions.}
    \label{fig:lfm_recons}
\end{figure*}

\begin{figure*}[!th]
    \centering
    \captionsetup[subfigure]{justification=centering}
    \begin{subfigure}[t]{0.25\textwidth}
        \includegraphics[width=\linewidth,trim={20 20 20 20},clip]{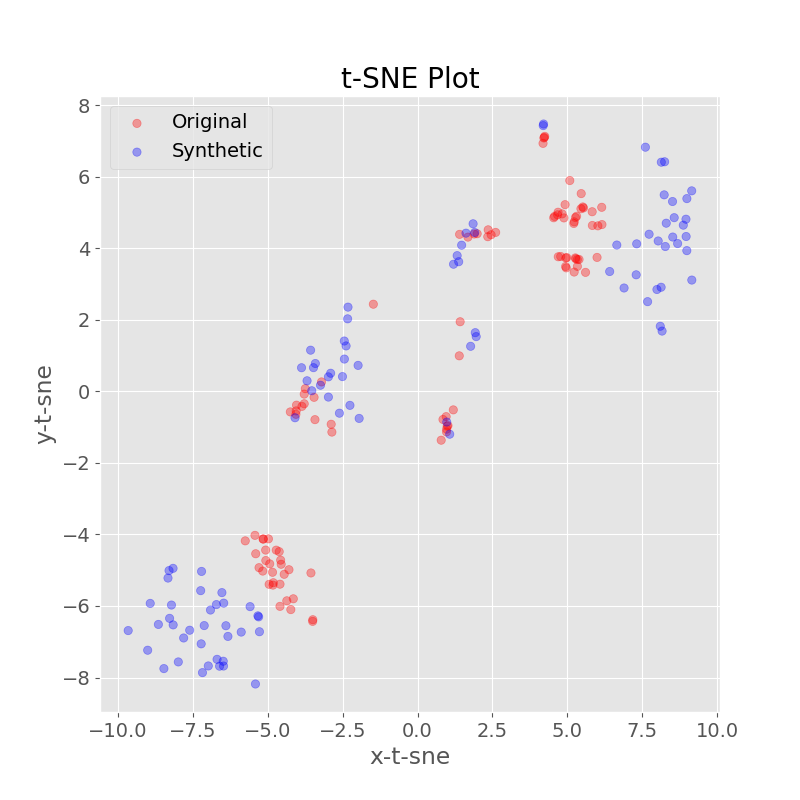}
        \caption{Baseline}
    \end{subfigure}\hfill
    \begin{subfigure}[t]{0.25\textwidth}
        \includegraphics[width=\linewidth,trim={20 20 20 20},clip]{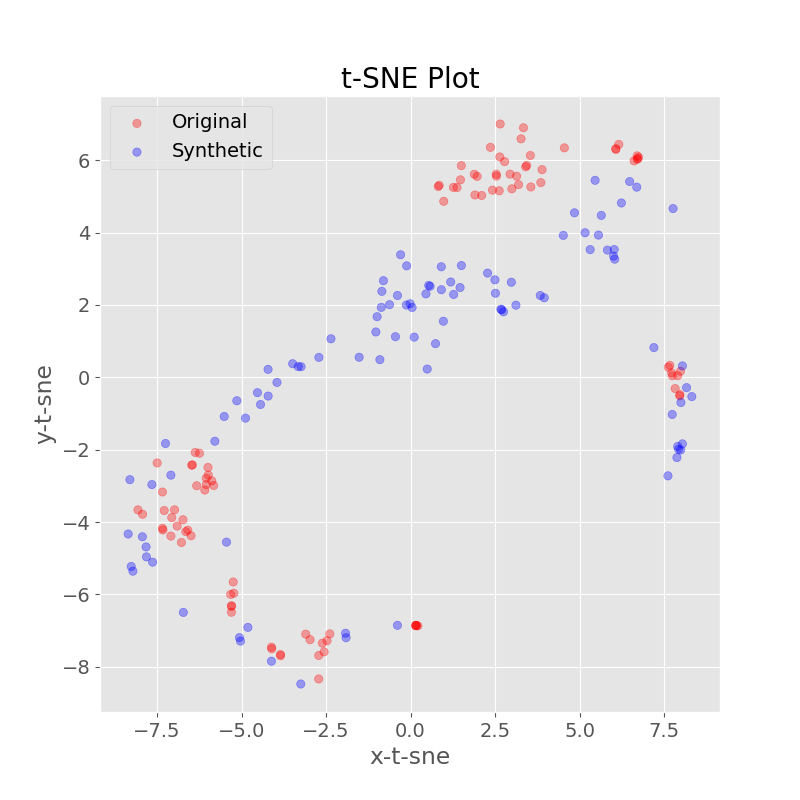}
        \caption{5\% split}
    \end{subfigure}\hfill
    \begin{subfigure}[t]{0.25\textwidth}
        \includegraphics[width=\linewidth,trim={20 20 20 20},clip]{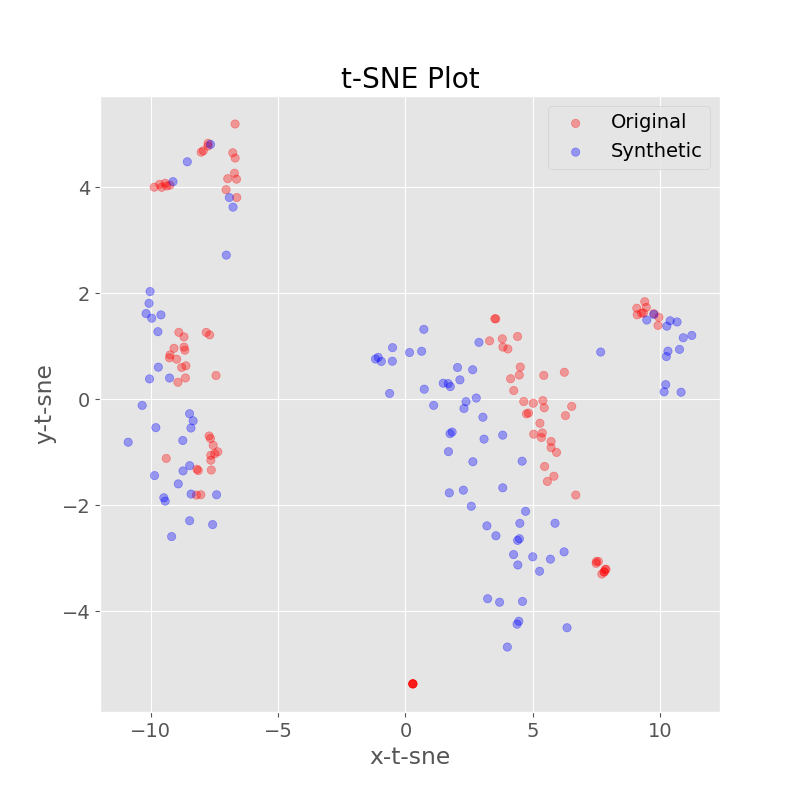}
        \caption{20\% split}
    \end{subfigure}\hfill
    \begin{subfigure}[t]{0.25\textwidth}
        \includegraphics[width=\linewidth,trim={20 20 20 20},clip]{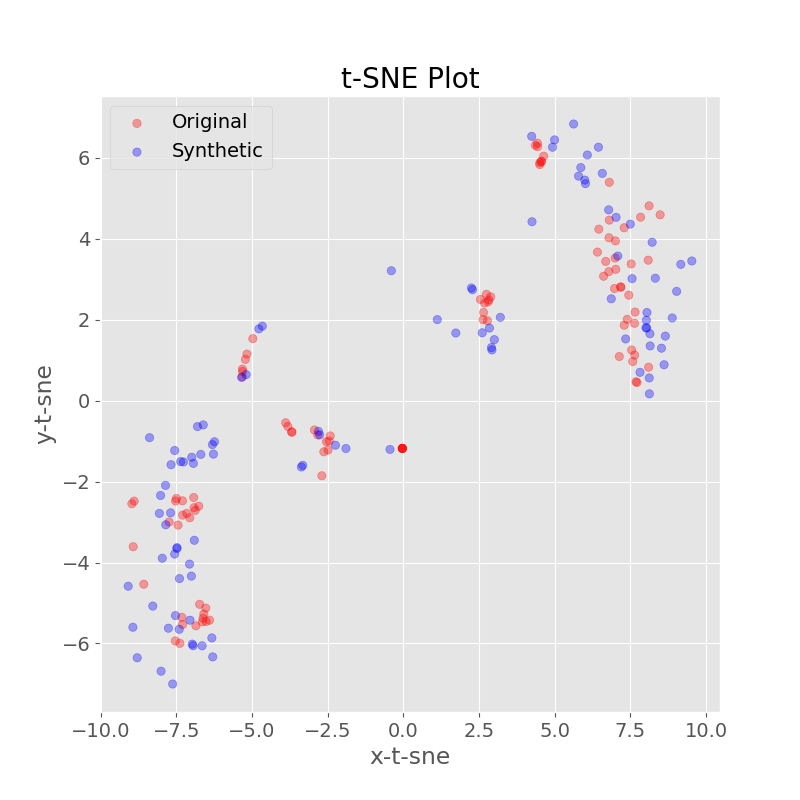}
        \caption{100\% split}
    \end{subfigure}
    \caption{LFM: t-SNE overlays (real in red, generated in blue) across Dublin data fractions.}
    \label{fig:lfm_tsne}
\end{figure*}

The trajectory overlays in \autoref{fig:lfm_recons} show a clear reduction of noise and spurious paths from $5\%$ through $100\%$, with realistic reproduction of the main eastern corridors and persistent difficulty in the rare west-facing approaches. The t-SNE visualizations in \autoref{fig:lfm_tsne} confirm the progression: poor baseline coverage, improved cluster overlap at $20\%$, and substantially better alignment at $100\%$, albeit with residual outliers. While LFM does not match DM’s sample efficiency, its late-stage performance becomes competitive once sufficient target data are available.

\begin{table}[!h]
\centering
\caption{LFM transfer results on Dublin (track/groundspeed). Lower is better.}
\label{tab:experiment:tl_track:lfm}
\begin{tabular}{lccc}
\toprule
\textbf{Split} & \textbf{e-distance} & \textbf{MMD} & \textbf{DTW} \\
\midrule
Baseline & 0.809 $\pm$ 0.212 & 0.168 $\pm$ 0.142 & 37.980 $\pm$ 0.116 \\
\midrule
0.00     & 0.775 $\pm$ 0.215 & 0.277 $\pm$ 0.167 & 48.915 $\pm$ 0.166 \\
0.05     & \textbf{0.732 $\pm$ 0.186} & 0.147 $\pm$ 0.142 & \textbf{35.604 $\pm$ 0.127} \\
0.20     & \textbf{0.705 $\pm$ 0.182} & \textbf{0.081 $\pm$ 0.151} & \textbf{32.555 $\pm$ 0.119} \\
0.50     & \textbf{0.704 $\pm$ 0.183} & 0.174 $\pm$ 0.182 & \textbf{34.462 $\pm$ 0.116} \\
1.00     & \textbf{0.677 $\pm$ 0.177} & \textbf{0.060 $\pm$ 0.160} & \textbf{27.433 $\pm$ 0.104} \\
\bottomrule
\end{tabular}
\end{table}

%\vspace{8pt}
\subsection*{Summary across models}
Pretraining consistently enhances \emph{sample efficiency} across architectures. DM shows the strongest transfer: with only $5$–$20\%$ target data it reaches baseline-level e-distance/MMD and \emph{significantly} improves DTW. LDM benefits similarly, though gains are smaller. LFM begins from a weaker baseline but exhibits large relative improvements as more target data become available, culminating in strong DTW/MMD at $100\%$. FM benefits from transfer as well, but its gains are smaller and more variable; it achieves competitive DTW at larger data budgets while lagging on distributional coverage in some splits. Across models, dominant approach corridors are reproduced reliably after transfer, while rarer runway usages (e.g., west-facing arrivals) remain underrepresented—highlighting opportunities for targeted augmentation or curriculum-based fine-tuning to emphasize rare modes.

%% file: sections/conclusion.tex
\section{Conclusion}

This paper presented, to the best of our knowledge, the first systematic study of transfer learning for deep generative models of aircraft landing trajectories. We evaluated four generative model families on transfer from a data-rich source airport (Zürich, LSZH) to a data-scarce target airport (Dublin, EIDW). Our findings show that pretraining consistently improves sample efficiency across architectures. Diffusion-based models exhibited the strongest transfer, achieving competitive performance with as little as 5\% of the target data and matching or surpassing full-data baselines around 20\%. Latent variants also benefited from pretraining: LDM followed the same trend as DM with smaller effect sizes, while LFM, despite its weaker baseline, gained disproportionately from additional target data and became competitive at larger splits. FM models also benefited from transfer, though their improvements were smaller and more variable.  Across all models, transfer learning enabled realistic reproduction of dominant approach corridors while rare modes, such as west-facing arrivals at Dublin, remained underrepresented. These results demonstrate that generative knowledge can be effectively transferred between airports, reducing the amount of local data required for high-quality trajectory synthesis. Future work should explore strategies to better capture rare trajectory patterns, including targeted augmentation, curriculum-based fine-tuning, and multimodal conditioning (e.g., weather). By establishing transfer learning as a viable approach, this study broadens the applicability of synthetic data generation in ATM, supporting scalable data-driven solutions even in operational contexts where historical records are limited.